%% file: main.tex
\PassOptionsToPackage{table}{xcolor}
\documentclass[11pt]{article}

\usepackage{acl}

\usepackage{times}
\usepackage{latexsym}
\usepackage{amsmath}
\usepackage{enumitem}
\usepackage{amsfonts}
\usepackage{booktabs}
\usepackage{adjustbox}
\usepackage{multirow}
\usepackage[table]{xcolor}
\usepackage[most]{tcolorbox}

\usepackage[T1]{fontenc}

\usepackage{microtype}

\usepackage{inconsolata}

\usepackage{graphicx}

\usepackage{xspace}

\definecolor{bl}{rgb}{0,0.2,0.6}
\definecolor{darkgreen}{rgb}{0,0.5,0}
\definecolor{red}{rgb}{1,0,0}
\definecolor{brown}{rgb}{0.6,0.3,0}
\definecolor{orange}{rgb}{1,0.5,0}
\title{\projectname{}:    Reasoning-aware Dense Retrieval Models}

\author{
 \textbf{Debrup Das\textsuperscript{1}},
  \textbf{Sam O’Nuallain\textsuperscript{1}},
\textbf{Razieh Rahimi\textsuperscript{1}}
\\
\\
 \textsuperscript{1}University of Massachusetts Amherst
\\ \\
 {
  debrupdas@umass.edu, sonuallain@umass.edu, rahimi@cs.umass.edu 
 }
}

\newcommand{\projectname}{RaDeR\xspace}

\begin{document}
\maketitle
\begin{abstract}
We propose \projectname, a set of reasoning-based dense retrieval models trained with data derived from mathematical problem solving using large language models (LLMs). Our method  leverages retrieval-augmented reasoning trajectories of an LLM and self-reflective relevance evaluation, enabling the creation of both diverse and hard-negative samples for reasoning-intensive relevance.
\projectname{} retrievers, trained for mathematical reasoning, effectively generalize to diverse reasoning tasks in the  \textsc{Bright} and RAR-b benchmarks, consistently outperforming strong baselines in overall performance. Notably, \projectname{} achieves significantly higher performance than baselines on the \emph{Math} and \emph{Coding} splits.
In addition, \projectname{} presents the first dense retriever that outperforms  BM25 when queries are Chain-of-Thought reasoning steps, underscoring the critical role of reasoning-based retrieval to augment reasoning language models. 
Furthermore, \projectname{} achieves comparable or superior performance 
while using only 2.5\% of the training data used by the concurrent work \textsc{ReasonIR}, highlighting the quality of our synthesized training data. 
Our code,  data, and retrieval models are publicly available.\footnote{\url{ https://anonymous.4open.science/r/project-D27D/}}
\end{abstract}

\input{intro2}

\input{related_work}

\input{task}

\input{IR4Math}

\input{exp}

\input{results}
\input{conclusion}
\input{limitation}

\bibliography{references}
\clearpage
\appendix
\input{Appendix_A}

\input{Appendix_B}
\end{document}

%% file: intro2.tex
\section{Introduction}
Large language models (LLMs) have demonstrated impressive reasoning capabilities on a wide range of tasks. 
Yet, they often benefit from  retrieval augmentation to enhance  accuracy, attributability~\cite{DBLP:journals/corr/abs-2403-03187}, and the interpretability~\cite{nakano2022webgptbrowserassistedquestionansweringhuman} of their outputs.
Retrieval models for LLM augmentation generally perform reasonably well at  lexical and semantic term matching, however they face challenges when reasoning is needed for relevance prediction~\cite{su2024brightrealisticchallengingbenchmark}.

\begin{figure}[t]
  \includegraphics[scale=0.46]{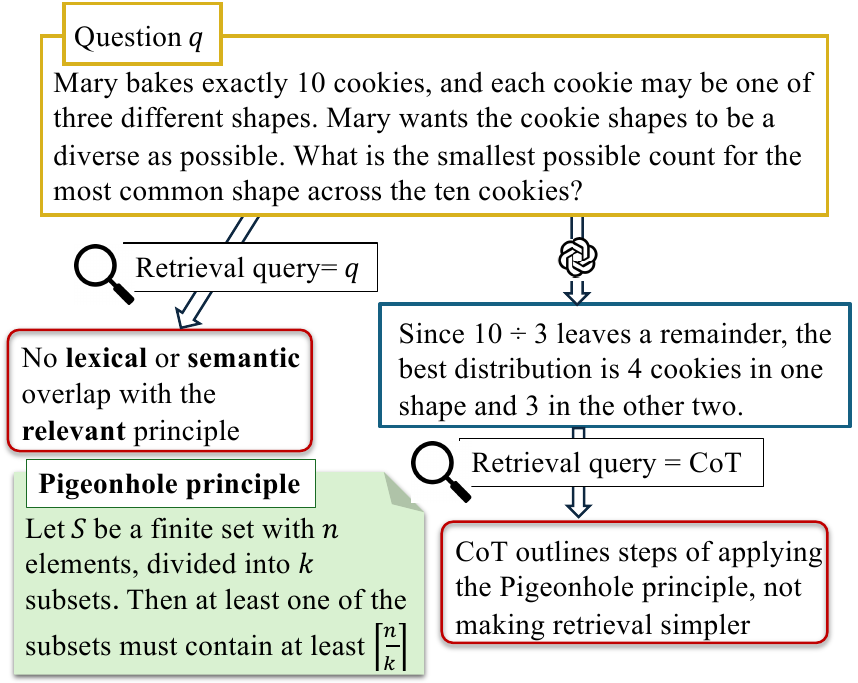}
  \caption{An example based on sample `TheoremQA\_jianyuxu/pigeonhole3' of \textsc{Bright}, where term matching retrievers face challenges in retrieving the relevant theorem w.r.t. both questions and CoT reasoning. }
  \label{fig:pigeonhole-example}
\end{figure}

Recent works have tried to address the reasoning limitation of existing models for relevance prediction. 
Two main approaches have emerged: (1)~interleaved reasoning and retrieval~\cite{hu2025mctsragenhancingretrievalaugmentedgeneration,jin2025searchr1trainingllmsreason,song2025r1searcherincentivizingsearchcapability}, and (2)~reasoning-based re-ranking models~\cite{weller2025rank1testtimecomputereranking,samarinas2025distillationrefinementreasoningsmall}. 
 While the first group is more effective than in-context retrieval augmentation, they are limited to the reasoning steps of LLMs for retrieval. As the example in Figure~\ref{fig:pigeonhole-example} shows, the reasoning steps of LLMs may not align with those needed for retrieval. 
 Solving the question in this example requires retrieving the \emph{pigeonhole principle}, where there are no matching terms between the question and the principle. 
 The reasoning steps by GPT-4 also do not simplify the retrieval of the pigeonhole principle, since they outline the steps of applying the pigeonhole principle to  solve the question.
On the other hand, reranking models are inherently limited by the candidate set produced by the first-stage retriever. 
To the best of our knowledge, there are \emph{no first-stage reasoning-based retrieval} models. 

Developing reasoning-based retrievers poses multiple \emph{challenges}.
The primary challenge is automatic generation of diverse and high-quality training data.
Specifically, training data should include queries of \emph{diverse formats and lengths} as well as samples with varying \emph{degrees of reasoning complexity}.
Beyond this, training retrieval models using representation learning~\cite{dai2023promptagator} presents an additional challenge due to the need for generating  hard-negative reasoning samples.

\begin{figure}[h]
\centering
\hspace{-0.2cm}\includegraphics[width=0.5\textwidth]{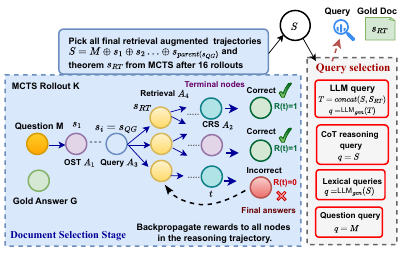}
\caption{An overview of the \textbf{\projectname} data generation pipeline. The \emph{OST} action stands for \textit{one step thought} generation, and  \emph{CRS} stands for \emph{complete remaining solution} steps action. 
} \label{fig:mainfigure}. 
\end{figure}

We propose \projectname, a set of first-stage reasoning-based retrieval models trained with synthesized data from mathematical reasoning. 
Specifically, we use an LLM for mathematical problem solving with a retrieval-augmented search-based reasoning approach, where the LLM can retrieve and apply theorems needed for solving intermediate subproblems.
To generate training data, we then sample reasoning trajectories with retrieval nodes 
based on the assumption that information retrieved during intermediate steps of the LLM’s search process is likely to be relevant to the original question.
This approach, illustrated in Figure~\ref{fig:mainfigure}, naturally yields a diverse set of queries, varying in length and complexity. 

In addition, verifying LLM-generated answers to mathematical questions provides a proxy for evaluating the relevance of retrieved information. To further ensure the quality of generated training data, the relevance evaluation is enhanced by \textit{self-reflection}. These evaluations help with generation of high-quality data. Additionally, any retrieved theorem evaluated as non-relevant is considered as a hard-negative reasoning sample.

\vspace{-5mm}We perform a comprehensive evaluation of \projectname{} models, including evaluation of retrieval performance on reasoning-intensive  benchmarks, traditional  benchmarks mainly requiring term matching, as well as evaluation of target QA performance using retrieval augmentation.  
Experimental results on \textsc{Bright}~\cite{su2024brightrealisticchallengingbenchmark} show that \projectname{} outperforms strong baselines by at least \textbf{2} points in different settings. These findings demonstrate that training retrieval models for mathematical reasoning effectively generalizes to other types of reasoning for information retrieval. In addition to improvements in overall performance, \projectname{} demonstrates particularly strong performance in \emph{Math} and \emph{Coding} splits of \textsc{Bright}. We observe  nDCG@10 relative improvements of \textbf{37-40\%}  in  the theorem-Q split, and \textbf{8-26\%} over the Leet coding split. 
\projectname{} presents the \emph{first}  dense retriever that outperforms BM25 in the zero-shot setting of using reasoning steps as retrieval queries. This achievement provides strong evidence for the necessity of reasoning-based retrievers even when retrievers augment reasoning language models. 
On the MMTEB~\cite{enevoldsen2025mmteb} reasoning subset, RAR-b~\cite{xiao2024rarbreasoningretrievalbenchmark}, \projectname significantly outperforms all sparse and open-source models, performing on par with large proprietary models such as OpenAI-3-large.

In a concurrent work, \citet{shao2025reasonir} also train reasoning-based first-stage  retrieval models. \projectname{} achieves a \textbf{1.1} point increase in nDCG@10 performance,  corresponding to a \textbf{4.5}\% relative gain. Performance of \projectname{} is particularly significant given that it is trained with 43,120 samples, about  \textbf{2.5}\% of samples used for \textsc{ReasonIR} (1,729,368), demonstrating the \emph{effectiveness} of \emph{our synthesized data}.

%% file: related_work.tex
\section{Related Works}

\textbf{Interleaving reasoning with retrieval.} 
Previous works have explored interleaving chain-of-thought \textbf{(CoT) reasoning} with retrieval using \textit{off-the-shelf retrievers}~\cite{trivedi2023interleavingretrievalchainofthoughtreasoning,shao-etal-2023-enhancing,yao2023react,schick2023toolformer} (see Appendix \ref{subsec:cotreasoning} for details). In this paradigm, a \textit{retrieval action} at a given step leverages a subset of the previously generated  CoT reasoning steps as the \textit{query} for retrieval. However, these methods are limited by their reliance on off-the-shelf retrievers.

Recent works like Search-R1~\cite{jin2025searchr1trainingllmsreason} and R1-Searcher~\cite{song2025r1searcherincentivizingsearchcapability} use reinforcement learning to optimize \textbf{reasoning-based query generation} while keeping the retrieval model fixed. In complex retrieval tasks, direct lexical or  semantic overlap between the initial question and the relevant  document is often limited. By incorporating intermediate CoT reasoning during retrieval, the system can better bridge this gap. Recent research explores \textbf{search-based methods} for exploring the space of CoT reasoning paths, such as random sampling \cite{wang2023selfconsistency} and Monte Carlo Tree Search (MCTS)~\cite{qi2025mutual,yao2023tree,guan2025rstarmathsmallllmsmaster,hao2023reasoning}. Retrieval-augmented generation (RAG) has also been integrated into MCTS-based frameworks by introducing retrieval-related \textit{actions} such as query generation and document retrieval~\cite{tran2024rareretrievalaugmentedreasoningenhancement, hu2025mctsragenhancingretrievalaugmentedgeneration}.

\textbf{Reasoning-based re-rankers.}
Recent methods such as \textsc{Rank-1}~\cite{weller2025rank1testtimecomputereranking} and InteRank~\cite{samarinas2025distillationrefinementreasoningsmall} train re-ranking models using knowledge distillation from reasoning LLMs and reinforcement learning respectively (see Appendix \ref{subsec:reasoningrerank} for details). However, these approaches remain limited by the candidate set of initial retrievers.

\textbf{Data augmentation for IR.}
Existing methods~\cite{nogueira-etal-2020-document, bonifacio2022inparsdataaugmentationinformation, dai2023promptagator} expand queries or documents with likely terms and generate new queries from existing documents. Recent methods~\cite{hu2024sertsselfrewardingtreesearch, lee2024geckoversatiletextembeddings} leverage LLMs to build iterative pipelines for synthetic data generation - using LLM-as-judge as the primary signal for data quality. The concurrent work ReasonIR~\cite{shao2025reasonir} also targets reasoning-intensive search, but generates queries without labeled scalar rewards.

Our work is the first to generate synthetic datasets specifically tailored for \textit{reasoning-intensive information retrieval}. Unlike previous methods, \projectname directly integrates reasoning into first-stage retrieval, leveraging MCTS to create sample-efficient synthetic data.

%% file: task.tex
\section{\projectname: Reasoning-aware Retrievers}

We propose a framework  that includes  a  first-stage retriever and  a re-ranking model, both performing reasoning to predict relevance. 
For the \textbf{first-stage retriever}, we adopt a uni-embedding bi-encoder architecture of dense retrieval models~\cite{lei-etal-2023-unsupervised}. 
For \textbf{re-ranking}, 
we fine-tune a pointwise \textit{cross-attention} model that takes the concatenation of the query and document. 
Our re-ranker directly predicts relevance scores. In contrast, existing reasoning-based re-rankers~\cite{weller2025rank1testtimecomputereranking, samarinas2025distillationrefinementreasoningsmall} rely on test-time compute for reasoning, making our approach significantly more  efficient at inference time.

%% file: IR4Math.tex
\section{Generating Retrieval Training Data}

The main challenge of developing reasoning-based retrieval models is synthesizing effective data that includes queries of \emph{diverse formats and lengths}, requiring varying degrees of reasoning complexity.  This diversity is essential for adaptive RAG systems~\cite{asai2023selfraglearningretrievegenerate, hu2025mctsragenhancingretrievalaugmentedgeneration, jin2025searchr1trainingllmsreason}, where LLMs may call retrievers at any intermediate reasoning or solution step.

A widely used data augmentation technique in IR involves prompting LLMs with passages to generate relevant queries~\cite{dai2023promptagator,bonifacio2022inparsdataaugmentationinformation, lee2024geckoversatiletextembeddings, hu2024sertsselfrewardingtreesearch}. 
However, this technique can limit the diversity of generated queries. It also poses challenges in verifying that the generated queries are relevant to the given passages through reasoning and are not generic, especially in the absence of ground truth for either the relevance reasoning steps or the queries.

To address the aforementioned challenges, we propose to generate training data using a retrieval-augmented Monte Carlo Tree Search (MCTS) reasoning approach~\cite{10.1007/11871842_29}  to solve mathematical problems by LLMs. Our motivation is twofold. 
First, solving mathematical problems often requires applying theorems to subproblems, which enables the integration of retrievers. Theorems found to be relevant to subproblems are also relevant to the original question due to the reasoning steps that connect them. Second, verifying LLM answers against gold answers provides a proxy for evaluating the utility of retrieved theorems in solving subproblems.

 \subsection{Retrieval-Augmented Search-based Reasoning} 
\label{sec:4.2}
We use a framework for solving mathematical problems by LLMs using a Monte Carlo Tree Search (MCTS) process, augmented with  retrieval over a collection of theorems and guided by scalar feedback based on the gold answers.

\noindent \textbf{MCTS overview.}
To solve a math problem $M$ from a given dataset, the MCTS algorithm prompts an LLM denoted by $\mathrm{LLM}_{\mathrm{gen}}$ to incrementally build a search tree that explores possible reasoning trajectories toward the final answer.
The generation process in MCTS is driven by two basic components: an \textit{action space} $A$ and a \textit{reward function} $R$. 
The root node of the tree corresponds to the input question $M$.
An edge from each node represents an action $a_i \in A$, and the resulting child node is an intermediate reasoning step $s_i$, which  is generated by applying action $a_i$ to the current reasoning trajectory $M \oplus s_1 \oplus s_2 \oplus \cdots \oplus s_{i-1}$.
Each node in the tree is assigned a value $Q(s,a)$, which represents the expected \textit{reward} on taking action $a$ from node~$s$. Initially, all nodes have $Q(s,a)=0$, resulting in a random tree exploration. As the algorithm performs rollouts, the $Q$ values of the nodes are updated based on  the rewards $R$ and the search is guided toward better reasoning trajectories.

\textbf{Action space.}
We extend the rStar framework~\cite{qi2025mutual} by adding two new actions for query generation and retrieval. As part of the \textit{retrieval action}, \projectname performs two additional steps:  \textit{self-reflection} which evaluates the relevance of retrieved theorems to the current problem context, and \textit{self-summarization} which generates concise natural-language summaries of the theorems to facilitate their integration into subsequent reasoning steps. The action space $A$ is described in detail below:

\noindent \textbf{$A_1$: Propose One-Step Thought (OST).} This action uses $\mathrm{LLM}_{\mathrm{gen}}$ to generate a single CoT reasoning step $s_i$ based on the context $M \oplus s_1 \oplus s_2 \oplus \cdots \oplus s_{i-1}$. The result node can be a \textit{terminal} node if $s_i$ contains the answer in \textit{boxed} notation (prompts in Appendix~\ref{OST:appendix}).

\noindent \textbf{$A_2$: Propose Complete Reasoning Steps (CRS).} This action uses $\mathrm{LLM}_{\mathrm{gen}}$ to complete the full solution by generating $s_i$ containing multiple reasoning steps, based on the current context $M \oplus s_1 \oplus s_2 \oplus \cdots \oplus s_{i-1}$. The result node is always a \textit{terminal} node of the MCTS. Prompt details are provided in Appendix~\ref{CRS:appendix}. 

\noindent \textbf{$A_3$: Generate Query (QG).}
This action prompts $\mathrm{LLM}_{\mathrm{gen}}$ to generate a retrieval query node $s_{\mathrm{QG}}$ based on the current context $M \oplus s_1 \oplus s_2 \oplus \cdots \oplus s_{\mathrm{parent(QG)}}$. A \textit{generate query} action is always followed by a \textit{retrieve theorem} action. For action $A_3$, we use few-shot prompting, details in Appendix~\ref{querygen:appendix}.

\noindent \textbf{$A_4$: Retrieve Theorem (RT).}
This action uses a retriever to obtain the top-$k$ theorems $s_{\mathrm{RT}}$ from the theorem corpus using the query $s_{\mathrm{QG}}$ of its parent node.
Each of the $k$ retrieved theorems that pass self-reflection is then added as a child node $s_{\mathrm{RT}}$.

\noindent \textbf{Self-Reflection and summarization.} 
To avoid expanding the search tree with irrelevant theorem nodes, we adopt \textit{self-reflection}~\cite{asai2023selfraglearningretrievegenerate,Xia_Zhou_Shi_Chen_Huang_2025} to evaluate the relevance of retrieved theorems.
We use few-shot prompting to guide $\mathrm{LLM}_{\mathrm{gen}}$ in generating both a relevance label (``relevant'' or ``non-relevant'') and a supporting explanation. The input comprises the original math question $M$, the current intermediate solution path ($M \oplus s_1 \oplus s_2 \oplus \cdots \oplus s_{\mathrm{parent(QG)}}$), and the retrieved theorem $s_{\mathrm{RT}}$.
Only theorems labeled as relevant are added as nodes in the MCTS tree, while non-relevant ones are pruned early to reduce unnecessary expansion and computation. This mechanism enables \projectname to focus exploration on more promising retrieval-augmented solution trajectories. We provide the prompts used for \textit{self-reflection} and \textit{self-summarization} in Appendix~\ref{selfreflection:appendix}. 

\noindent \textbf{Sampling solutions with MCTS rollouts.}
We sample multiple candidate solution trajectories with MCTS rollouts following  rStar~\cite{qi2025mutual}.  The details  are described in Appendix~\ref{sec:MCTSrollouts}. 


The \textbf{reward function} $R(t)$ for a terminal node $t$ is calculated based on whether the solution trajectory reaches the correct answer to the input question $M$. Specifically, if the trajectory leads to the correct answer, $R(t)=1$; otherwise, $R(t)=0$.

\subsection{Synthesizing Training Data}
To build training data for reasoning-based retrievers, we sample high-reward solution trajectories generated by the MCTS framework. 
For a math question $M$, we extract all  solution trajectories that contain at least one retrieval node, denoted as \( S = M \oplus s_1 \oplus s_2 \oplus \cdots \oplus s_{\mathrm{QG}} \oplus s_{\mathrm{RT}} \oplus \cdots \oplus s_t \).  
From each selected solution trajectory $S$, we generate training samples in the form $(q, p, N)$, where $q$ denotes a query, $p$ is 
 a positive theorem relevant to $q$, and $N$ is a set of hard-negative theorems for reasoning-intensive retrieval.
The retrieved theorem \( s_{\mathrm{RT}} \) in $S$ is used as the positive document $p$ of the generated samples. 
Any theorem retrieved with respect to $s_{\mathrm{QG}}$ that was not labeled as relevant in self-reflection is included in the set of hard negatives $N$.

\noindent\textbf{Reasoning-intensive data.}
To generate reasoning samples, the set of positive and hard-negative theorems is paired with three types of queries. 
(1)~\emph{MCTS CoT reasoning queries:} we use the CoT reasoning steps (partial solution) up to the query node, $M \oplus s_1 \oplus s_2 \oplus \cdots \oplus s_{\mathrm{parent(QG)}}$, as the retrieval query. 
Note that we do not use the  query $s_{\mathrm{QG}}$ generated by MCTS for retrieval training, since the specific prompt format for their generation may limit the diversity of our training data. 
These queries are denoted as $q_{\mathrm{CoT}}$. 
(2)~\emph{LLM reasoning queries:} We prompt the $\mathrm{LLM}_{\mathrm{gen}}$ to generate a query based on the math question $M$, the reasoning steps up to the query node (excluding the query), and the theorem $s_{\mathrm{RT}}$. This group of queries is referred to as $q_{\mathrm{llmq}}$. 
Using few-shot prompting, the $\mathrm{LLM}_{\mathrm{gen}}$ is guided to generate reasoning-intensive queries that have low lexical and semantic term overlap with the positive theorems. Unlike the query generated during the QG action of MCTS, which directs $\mathrm{LLM}{_\mathrm{gen}}$ to produce a hypothetical theorem for the current subproblem, $q_{\mathrm{llmq}}$ is a concise query that directly reflects the information need of the subproblem. The prompt template for this query is provided in Appendix~\ref{LLMquery:appendix}. 
(3)~\emph{Questions as queries}: the input math question $M$ is used as the query for  retrieving theorem $s_{\mathrm{RT}}$. Given the low lexical and semantic overlap between $M$ and $s_{\mathrm{RT}}$, it serves as a good reasoning sample for training. We denote this type of queries as $q_{\mathrm{question}}$.

\textbf{Term-matching data.}
Training a reasoning-based retriever should not hurt the performance of queries that can be addressed with lexical/semantic term matching. To achieve this, we follow the widely used approach of synthetically generating training samples for retrieval models. Specifically, to generate queries that have high term similarity with their respective relevant theorems, we prompt LLM$_\mathrm{gen}$ using only the theorem $s_{\mathrm{RT}}$. Following Promptagator~\cite{dai2023promptagator}, we add a filtering step based on \textit{round-trip consistency}. 
Only the queries for which BM25 retrieves the corresponding positive theorem $s_{\mathrm{RT}}$ within its top-$k$ ($k$=20) results are kept.
This filtering step typically results in queries with large term overlap. Hard negatives are extracted from top results of BM25 and RepLLaMA retrieval models.
We denote this type of queries as $q_{\mathrm{lexical}}$.\\
\textbf{Synthesized data mix.}
We construct our retrieval training dataset by combining all query types, including both reasoning-based and lexical queries. Detailed statistics of the synthesized training data are provided in Appendix~\ref{sec:stats-training-data}.

%% file: exp.tex
\section{Experimental Settings}

\textbf{Base language models.} To train dense retrieval models, we primarily utilize the instruction-tuned variants of the \emph{Qwen2.5} suite of LLMs~\cite{yang2024qwen2}. 
For ablation studies on model size, we train a series of Qwen-2.5-instruct models with varying parameter sizes ranging from 3B, 7B to 14B. We additionally perform ablation studies using different LLMs including gte-Qwen2-7B-Instruct~\cite{li2023generaltextembeddingsmultistage} and Llama-3.1-Instruct~\cite{grattafiori2024llama}.

\textbf{Math datasets  for data generation.} 
We incorporate two datasets for mathematical problem solving in \projectname{}: MATH~\cite{hendrycks2021measuringmathematicalproblemsolving} and a subset of examples from {NuminaMath}~\cite{numina_math_datasets}. Details are provided in  Appendix~\ref{details:mathdataset}.

\input{Tables/table-bright-retrieval-questions}
\input{Tables/table-bright-retrieval-cot}

\textbf{Retrieval-augmented MCTS.} We apply the \projectname{} framework using a fixed set of parameters, as detailed in Appendix~\ref{mctsparams}. For each input math question in a dataset for mathematical problem solving, we perform 16 rollouts.
We use RepLLaMA~\cite{ma2023finetuningllamamultistagetext} as the retriever in our MCTS algorithm.
The retrieval corpus consists of formal mathematical theorems from {ProofWiki},\footnote{\url{https://proofwiki.org/} — a comprehensive collection of over 20K formal definitions and theorem proofs} which is also used in the \textsc{Bright} benchmark~\cite{su2024brightrealisticchallengingbenchmark}.

\textbf{Training details.}
We train both the \textit{retriever} and the \textit{re-ranker} for reasoning-intensive  relevance prediction. The retriever is trained using the standard contrastive \textit{InfoNCE} loss using in-batch negatives in addition to the hard negatives from synthesized training samples. We use 12 hard negatives per query and treat passages from other examples in the batch as in-batch negatives.
The retriever training details are presented in Appendix~\ref{sec:appendix_a_trainingloss}.  
Similar to the retriever, the reranker is trained using contrastive loss. The reranker training details are presented in Appendix~\ref{sec:appendix_a_trainingloss_reranker}.The hyperparameters used for training are presented in Table~\ref{table:ranking-hyper-params}.

%% file: Tables/table-bright-retrieval-questions.tex
\begin{table*}[htb]
\centering
\scriptsize
\resizebox{0.97\textwidth}{!}
{\begin{tabular}{l|ccccccc|cc|ccc|c}
\hline
& \multicolumn{7}{c|}{StackExchange} & \multicolumn{2}{c|}{Coding} & \multicolumn{3}{c|}{Theorem-based} & \multirow{2}{*}{\textbf{Avg.}} \\
\cline{2-8} \cline{9-10} \cline{11-13}
& Bio & Earth & Econ & Psy & Rob & Stack & Sus & Leet & Pony & AoPS & TheoQ & TheoT & \\
\hline
\rowcolor{brown!20}
\multicolumn{14}{c}{Sparse and Open-source Baselines} \\
\hline
BM25$^\star$& 18.9 & 27.2 & 14.9 & 12.5 & 13.6 & 18.4 & 15.0 & 24.4 & 7.9 & 6.2 & 10.4 & 4.9 & 14.5 \\
Qwen$^\star$ & 30.6 & 36.4 & 17.8 & 24.6 & 13.2 & \underline{22.2} & 14.8 & 25.5 & 9.9 & \underline{14.4} & 27.8 & 32.9 & 22.5  \\
Qwen2 & \underline{34.1} & \textbf{42.6} & 18.2 & 27.4 & 13.2 & 17.3 & \underline{20.9} & 30.4 & 2.2 & 13.3 & 30.6 & 32.6 & 23.5  \\
GritLM$^\star$& 24.8 & 32.3 & 18.9 & 19.8 & \underline{17.1} & 13.6 & 17.8 & 29.9 & \textbf{22.0} & 8.8 & 25.2 & 21.2 & 21.0 \\
Inst-XL$^\star$& 21.6 & 34.3 & \textbf{22.4} & 27.4 & \textbf{18.2} & 21.2 & 19.1 & 27.5 & 5.0 & 8.5 & 15.6 & 5.9 & 18.9 \\
E5$^\star$ & 18.6 & 26.0 & 15.5 & 15.8 & 16.3 & 11.2 & 18.1 & 28.7 & 4.9 & 7.1 & 26.1 & 26.8 & 17.9 \\
\hline
\rowcolor{red!20}
\multicolumn{14}{c}{Proprietary Baselines} \\
\hline
Google$^\star$ & 22.7 & 34.8 & 19.6 & 27.8 & 15.7 & 20.1 & 17.1 & 29.6 & 3.6 & 9.3 & 23.8 & 15.9 & 20.0 \\
Voyage$^\star$& 23.1 & 25.4 & 19.9 & 24.9 & 10.8 & 16.3 & 15.4 & 30.6 & 1.5 & 7.5 & 27.4 & 11.6 & 17.9  \\
\hline
\rowcolor{orange!30}
\multicolumn{14}{c}{\projectname{} Models (MATH dataset, ($q_\mathrm{llmq} + q_\mathrm{CoT} + q_\mathrm{lexical}$) )} \\
\hline
Qwen2.5-7B-instruct & 25.4 & 30.0 & 16.7 & 25.3 & 14.0 & 21.3 & 16.3 & 37.0 & 8.2 & \textbf{15.7} & \textbf{42.7} & \underline{44.4} & \underline{24.6} \\
gte-Qwen2-7B  & \textbf{34.6} & \underline{38.9} & \underline{22.1} & \textbf{33.0} & 14.8 & \textbf{22.5} & \textbf{23.7} & \underline{37.3} & 5.0 & 10.2 & 28.4 & 35.1 & \textbf{25.5}  \\
Llama3.1-8B-Instruct  & 29.3 & 27.3 & 17.5 & \underline{28.2} & 12.1 & 18.2 & 16.1 & \textbf{38.6} & \underline{11.8} & 6.4 & 32.3 & 33.1 & 22.6  \\
\hline
\rowcolor{orange!30}
\multicolumn{14}{c}{\projectname Models (MATH$+$NuminaMath datasets , all query types)} \\ \hline
 Qwen2.5-7B-instruct  & 25.1 & 28.3 & 18.2 & 25.9 & 15.3 & 22.2 & 16.3 & 35.9 & 6.9 & 10.4 & \underline{40.8} & \textbf{47.1} & 24.4 \\
\hline
\end{tabular}
}
\caption{nDCG@10 performance of strong baselines and our models over the \textsc{Bright} benchmark using the original question as queries for retrieval. Results of models with $^\star$ are taken from~\cite{su2024brightrealisticchallengingbenchmark}.}
\label{table:retrieval-bright-original-questions}
\end{table*}

%% file: Tables/table-bright-retrieval-cot.tex
\begin{table*}[!htb]
\centering
\scriptsize
\resizebox{0.97\textwidth}{!}{
\begin{tabular}{l|ccccccc|cc|ccc|c}
\hline
& \multicolumn{7}{c|}{StackExchange} & \multicolumn{2}{c|}{Coding} & \multicolumn{3}{c|}{Theorem-based} & \multirow{2}{*}{\textbf{Avg.}} \\
\cline{2-8} \cline{9-10} \cline{11-13}
& Bio & Earth & Econ & Psy & Rob & Stack & Sus & Leet & Pony & AoPS & TheoQ & TheoT & \\
\hline
\rowcolor{brown!20}
\multicolumn{14}{c}{Sparse and Open-source Baselines} \\
\hline
BM25$^\star$ & \textbf{53.6} & \textbf{54.1} & 24.3 & \underline{38.7} & \underline{18.9} & 27.7 & \underline{26.3} & 19.3 & 17.6 & 3.9 & 19.2 & 20.8 & 27.0 \\
Qwen$^\star$  & 35.5 & 43.1 & 24.3 & 33.4 & 15.4 & 22.9 & 23.9 & 25.4 & 5.2 & 4.6 & 28.7 & 34.6 & 24.8 \\
Qwen2  & 38.3 & 47.3 & 24.0 & 35.2 & 15.9 & 23.3 & 27.9 & 29.5 & 8.9 & 2.9 & 30.8 & 35.1 & 26.6 \\
GritLM$^\star$  & 33.3 & 39.1 & 22.4 & 28.9 & 17.4 & 21.3 & 24.1 & 31.9 & 12.0 & 6.7 & 27.3 & 30.1 & 24.5 \\
Inst-XL$^\star$  & \underline{46.7} & \underline{51.2} & \textbf{29.9} & \textbf{40.5} & \textbf{20.8} & \textbf{30.1} & \textbf{26.9} & \textbf{35.1} & 2.1 & 8.2 & 24.2 & 17.0 & 26.9 \\
E5$^\star$  & 29.3 & 43.9 & 19.9 & 26.6 & 11.6 & 19.8 & 15.6 & 29.1 & 0.9 & 5.3 & 27.0 & 36.6 & 22.1 \\
\hline
\rowcolor{red!20}
\multicolumn{14}{c}{Proprietary Baselines} \\
\hline
Google$^\star$  & 36.4 & 45.6 & \underline{25.6} & 38.2 & 18.7 & \underline{29.5} & 15.7 & 31.1 & 3.7 & 10.0 & 27.8 & 30.4 & 26.2 \\
Voyage AI$^\star$  & 36.7 & 42.8 & 24.6 & 34.2 & 13.7 & 24.2 & 21.7 & 31.4 & 2.2 & 6.6 & 30.3 & 28.1 & 24.7 \\
\hline
\rowcolor{orange!30}
\multicolumn{14}{c}{\projectname{} Models (MATH dataset, $q_\mathrm{llmq} + q_\mathrm{CoT} + q_\mathrm{lexical}$)} \\
\hline
 Qwen2.5-7B-instruct  & 32.4 & 38.0 & 21.5 & 33.2 & 14.5 & 25.5 & 18.1 & 30.1 & 14.0 & \underline{11.4} & \textbf{42.1} & \textbf{47.2} & 27.3 \\
 gte-Qwen2-7B  & 36.1 & 42.9 & 25.2 & 37.9 & 16.6 & 27.4 & 25.0 & 34.8 & 11.9 & \textbf{12.0} & 37.7 & 43.4& \textbf{29.2} \\
 Llama3.1-8B-Instruct  &37.6 & 41.4 & 21.1 & 33.1 & 12.5 & 27.7 & 15.8 & \underline{35.0} & \textbf{23.6} & 7.1 & 36.9 & 40.5 & \underline{27.7} \\
\hline
\rowcolor{orange!30}
\multicolumn{14}{c}{\projectname{} Models (MATH$+$NuminaMath datasets, all query types)} \\ \hline
 Qwen2.5-7B-instruct  & 37.5 & 40.0 & 19.3 & 31.1 & 14.1 & 25.8 & 17.6 & 27.2 & \underline{18.7} & 9.9 & \underline{40.4} & \underline{43.8} & 27.4 \\
\hline
\end{tabular}
}
\caption{nDCG@10 performance using GPT-4 CoT reasoning as queries for retrieval over \textsc{Bright}. Results of models with $^\star$ are taken from~\cite{su2024brightrealisticchallengingbenchmark}.}
\label{table:retrieval-bright-CoT-queries}
\end{table*}

%% file: results.tex
\section{Experimental Results}

We comprehensively evaluate our \projectname{} retriever and re-ranker models on (1)~the reasoning-intensive benchmark \textsc{Bright}~\cite{su2024brightrealisticchallengingbenchmark} and MMTEB reasoning tasks based on RAR-b~\cite{xiao2024rarbreasoningretrievalbenchmark}, and (2)~widely used benchmark MS MARCO~\cite{bajaj2018msmarcohumangenerated} to measure term matching performance. We also evaluate the performance of reasoning LLMs when augmented with our \projectname{} models. 
Following previous studies~\cite{su2024brightrealisticchallengingbenchmark}, nDCG at top-10 (nDCG@10) is used as the evaluation metric to compare different models. We also report the recall and precision  of retrieval models in Appendix~\ref{prec-rec-on-bright}.

\subsection{ \textsc{Bright} Retrieval Performance } 

We first present the performance of our \projectname retrieval models on the \textsc{Bright} benchmark~\cite{su2024brightrealisticchallengingbenchmark}. 
Following the benchmark, we compare our models against a diverse set of baselines including (1) BM25~\cite{robertson1995okapi}, (2) open-source dense retrieval models Instructor-XL~\cite{su-etal-2023-one}, E5-Mistral~\cite{wang-etal-2024-improving-text}, GritLM~\cite{muennighoff2024generative}, gte-Qwen1.5~\cite{li2023generaltextembeddingsmultistage}, gte-Qwen2~\cite{li2023generaltextembeddingsmultistage}, and (3) proprietary models from Voyage~\cite{voyage} and Google~\cite{lee2024geckoversatiletextembeddings}.

Following the benchmark, all models are evaluated in two settings: (1)~using the original questions from \textsc{Bright} as  retrieval queries, and (2)~using CoT reasoning steps generated by GPT-4, included in the benchmark, as  retrieval queries. Performance results for the two settings are reported in Tables~\ref{table:retrieval-bright-original-questions}~and~\ref{table:retrieval-bright-CoT-queries}, respectively.

\textbf{Overall performance.}
 \projectname{}-gte-Qwen2-7B achieves the best average performance of \textbf{25.5} on \textsc{Bright}, outperforming strong baselines by at least \textbf{2} points in both query settings. These results demonstrate that training retrieval models for mathematical reasoning \emph{generalizes} to other types of reasoning required for different retrieval tasks.

\input{Tables/table-retrieval-rarb}

\textbf{Theorem-based splits.} 
Compared to baselines,  \projectname{} achieves the largest improvements on the TheoT and TheoQ splits in both settings, as shown in Tables~\ref{table:retrieval-bright-original-questions}~and~\ref{table:retrieval-bright-CoT-queries}.
For example,  \projectname{} models improve the performance of TheoQ by \textbf{12.1} and \textbf{11.3} points when questions and CoT reasoning are used as queries, respectively. 
Although these splits of \textsc{Bright} primarily require mathematical reasoning, the performance improvements on TheoQ are particularly noteworthy since the retrieval collection for this split  consists of questions, a format not seen during the training of \projectname{} models. The strong performance of \projectname{} on this split demonstrates its potential for tasks such as demonstration selection in in-context learning, a task shown to have significant impacts on the performance of LLMs~\cite{rubin-etal-2022-learning}.

\textbf{Coding splits.}
Results in Table~\ref{table:retrieval-bright-original-questions} show that \projectname{} achieves the best performance on \emph{LeetCode}, surpassing the strong baselines by \textbf{8} points.
Results on \emph{Pony} in the CoT reasoning setting, Table~\ref{table:retrieval-bright-CoT-queries}, also show improvements of \projectname{} over the strongest baseline. 
These results are particularly significant, as they demonstrate that training our retrievers for mathematical reasoning yields substantial improvements in code retrieval,  despite not having any code-specific training data. Our qualitative analysis reveals that in most cases, coding problems rely on solving underlying mathematical subproblems. We believe that the ability to recognize this mathematical substructure enables \projectname{} models to  enhance retrieval effectiveness on these coding splits. A qualitative example is provided in Appendix~\ref{analysis_code}.

\input{Tables/table_traditionalbenchmarks}
\input{Tables/table-reranking}
\input{Tables/table-reranking-gpt4o}
\textbf{Ablation studies.} We investigated the performance of \projectname{} retrievers under two ablation settings: (1)~training with different subsets of query types in the  synthesized data, and (2)~using base LLM of different sizes as the retriever encoder. Results in Table~\ref{table:queryablations} ofAppendix~\ref{ablations_models}  highlight the \emph{complementary} role of the diverse query types in our synthesized training data. 
Additionally, Table~\ref{table:sizeablations} shows consistent performance improvements with scaling the size of the base LLM.

\subsection{Retrieval Performance on RAR-b} 

We evaluate the performance of  \projectname{}  on the reasoning-retrieval tasks of MMTEB~\cite{enevoldsen2025mmteb} which are based on 
the Math and Coding splits of the RAR-b~\cite{xiao2024rarbreasoningretrievalbenchmark}. As shown in Table~\ref{table:RAR-B}, our model  achieves performance comparable  to the strongest open-source baseline, outperforming on the MATH split. Furthermore, it demonstrates performance that is comparable to or surpasses that of top-performing closed-source models. These results once again highlight the strong generalizability of \projectname{} across diverse reasoning-based retrieval tasks.

\subsection{Retrieval on Traditional IR Benchmarks} 

Table~\ref{table:trad_bench} presents the performance of \projectname{} and strong baselines on the MS MARCO passage retrieval task~\cite{bajaj2018msmarcohumangenerated}. We evaluate on the official small subset of the MS MARCO development set, following RepLLaMA~\cite{ma2023finetuningllamamultistagetext}, as well as TREC-DL'19~\cite{craswell2020overview} and TREC-DL'20~\cite{craswell2021overview}. These test sets  primarily rely on term matching and do not require complex reasoning. 
Our model based on gte-Qwen2-7B trained with $\mathrm{llmq}$, $\mathrm{CoT}$, and $\mathrm{lexical}$ queries from MATH, demonstrates competitive performance with strong baselines. 
These results indicate that training our retrievers for reasoning-intensive tasks does not compromise their effectiveness on standard IR benchmarks where reasoning is not necessary.

\input{Tables/table-e2e}

\subsection{\textsc{Bright} Reranking Performance} 

We report the performance of our \projectname{} rerankers on the reasoning-intensive \textsc{Bright} benchmark. Table~\ref{table:rerankbm25} shows the performance of reranking  top-10 and top-100  of BM25 results using questions as queries. Our \projectname{} rerankers based on Qwen2.5  and  gte-Qwen2, both trained on  all query types from MATH, outperform GPT-4 in the top-100 reranking setting by \textbf{+2.5} and \textbf{+2.1} nDCG points, respectively. The most significant improvements are observed on the LeetCode and TheoremQA questions splits, with gains of \textbf{+20.3} and \textbf{+20.4} points, respectively. 

Table~\ref{tab:rerankperformance} presents the performance of reranking BM25 results using GPT-4o CoT reasoning, where the rerankers receive only the question as the query. We compare  \projectname{} rerankers against strong baselines including RankLLaMA~\cite{ma2023finetuningllamamultistagetext}, Mono-T5-3B, and \textsc{Rank1}~\cite{weller2025rank1testtimecomputereranking}. \textsc{Rank1}, the strongest baseline, is trained on 635K examples from  MS MARCO, and utilizes test-time compute to perform reasoning before relevance prediction.  In contrast, \projectname{} models are trained   with substantially fewer samples on mathematical reasoning, 43K from MATH and 78K from NuminaMATH,  yet they achieve highly competitive performance. 
In addition, \projectname{} models are significantly more computationally efficient since they only generate relevance scores.

\subsection{RAG Performance using \projectname{}}

Table~\ref{tab:accuracy_comparison} provides the answer accuracy of Qwen-2.5-7B-Instruct on the TheoremQA  split of \textsc{Bright} in two settings of retrieval augmentation. First setting is  in-context RAG  where the input question is used as the query to retrieval models. 
We also perform evaluation using the retrieval-augmented MCTS framework, employing majority voting as the strategy for answer selection.
In both settings, augmenting with the results of \projectname{} outperforms augmentation with RepLLaMA. These results demonstrate the impact of augmenting strong LLMs with reasoning-based retrievers. 

%% file: Tables/table-retrieval-rarb.tex
\begin{table}[!htb]
\scriptsize
\centering
\resizebox{!}{0.37\linewidth}{
\begin{tabular}{lcc}
\hline
Model & Math & Coding \\
\hline
\rowcolor{brown!20}
\multicolumn{3}{l}{Open-source Baselines} \\
\hline
Contriever (w/ Inst.) & 0.218 & 0.071 \\
all-mpnet-base-v2 (w/ Inst.) & 0.692 & 0.488\\
all-MiniLM-L6-v2 (w/ Inst.) & 0.624 & 0.423 \\
Dragon+ (w/ Inst.) & 0.362 & 0.128 \\
Instructor-XL (w/ Inst.) & 0.580 & 0.495 \\
bge-large (w/ Inst.) & 0.498 & 0.453 \\
E5-Mistral (w/ Inst.) & 0.740 & 0.785 \\
GritLM (w/ Inst.) & 0.824 & \underline{0.838} \\
\hline
\rowcolor{red!20}
\multicolumn{3}{l}{Proprietary Models} \\
\hline
Cohere-Embed-v3 (w/ Inst.) & 0.721 & 0.566 \\
OpenAI-ada-002 (w/ Inst.) & 0.673 & 0.824 \\
OpenAI-3-large (w/ Inst.) & \textbf{0.877} & \textbf{0.894} \\
\hline
\rowcolor{orange!30}
\multicolumn{3}{l}{\projectname{} model (MATH, all query types)} \\
\hline
gte-Qwen2-7B-instruct (w/Inst.) & \underline{0.852} & 0.835 \\
\hline
\end{tabular}
}
\caption{nDCG@10 performance of retrievers on the Math  and Coding splits of RAR-b. Performance of baselines are from~\cite{xiao2024rarbreasoningretrievalbenchmark}.}
\label{table:RAR-B}
\end{table}

%% file: Tables/table_traditionalbenchmarks.tex
\begin{table}[ht]
\centering
\scriptsize
\begin{tabular}{l|cc|c|c}
\hline
\multirow{2}{*}{\textbf{Model}} & \multicolumn{2}{c|}{DEV} & DL19 & DL20 \\
& MRR@10 & R@1k & nDCG@10 & nDCG@10 \\
\midrule
BM25  & 18.4 & 85.3 & 50.6 & 48.0 \\
ANCE  & 33.0 & 95.9 & 64.5 & 64.6 \\
 CoCondenser  & 38.2 & 98.4 & 71.7 & 68.4 \\
 TAS-B  & 34.0 & 97.5 & 71.2 & 69.3 \\
GTR-base  & 36.6 & 98.3 & - & - \\
GTR-XXL  & 38.8 & 99.0 & - & - \\
OpenAI Ada2  & 34.4 & 98.6 & 70.4 & 67.6 \\
bi-SimLM  & 39.1 & 98.6 & 69.8 & 69.2 \\
RepLLaMA & \textbf{41.2} & \textbf{99.4} & \textbf{74.3} & \textbf{72.1} \\
SimLM & 41.1 & 98.7 & 71.4 & 69.7 \\ 
 \hline
\projectname{}& 34.4 & 98.1 & 71.2 & \underline{70.7} \\
\hline
\end{tabular}
\caption{Performance of baselines and \projectname{}-gte-Qwen2-7B (trained with MATH)  on MS MARCO. Performance of baselines are from~\cite{ma2023finetuningllamamultistagetext,wang2023simlmpretrainingrepresentationbottleneck}.
}
\label{table:trad_bench}
\end{table}

%% file: Tables/table-reranking.tex
\begin{table*}[!th]
\scriptsize
    \centering
    \small
    \renewcommand{\arraystretch}{1.2}
     \begin{adjustbox}{max width=\textwidth}
     \scriptsize
    \begin{tabular}{l c | c c c c c c c | c c | c c c | c}
        \hline
        \multirow{2}{*}{\textbf{Reranker}} & \multirow{2}{*}{\textbf{top-k}} & \multicolumn{7}{c|}{StackExchange} & \multicolumn{2}{c|}{Coding} & \multicolumn{3}{c|}{Theorem-based} & \multirow{2}{*}{\textbf{Avg.}} \\
        \cline{3-9} \cline{10-11}\cline{12-14}
        & & Bio. & Earth. & Econ. & Psy. & Rob. & Stack. & Sus. & Leet. & Pony & AoPS & TheoQ. & TheoT. &  \\
        \hline
        None$^\star$ & - & 19.2 & 27.1 & 14.9 & 12.5 & 13.5 & 16.5 & 15.2 & \underline{24.4} & 7.9 & 6.2 & 9.8 & 4.8 & 14.3 \\
        \hline
        MiniLM$^\star$ & 100 & 8.5 & 18.9 & 6.0 & 5.4 & 7.6 & 7.9 & 8.9 & 15.0 & 11.3 & 6.1 & 3.6 & 0.5 & 8.3 \\
        \hline
        Gemini$^\star$ & 10 & 21.9 & 29.7 & 16.9 & 14.2 & 16.1 & 16.7 & 16.7 & \textbf{24.5} & 8.0 & 6.2 & 9.5 & 8.2 & 15.7 \\
        \hline
        GPT-4$^\star$ & 100 & \textbf{33.8} & \textbf{34.2} & 16.7 & \underline{27.0} & \textbf{22.3} & \textbf{27.7} & 11.1 & 3.4 & 15.6 & 1.2 & 2.0 & 8.6 & 17.0 \\
        \hline
        \rowcolor{orange!30}
        \multicolumn{15}{c}{\projectname{} models (MATH, all query types)} \\
        \hline
         Qwen2.5-7B-instruct  & 100 & \underline{26.9} & 30.6 &  17.0 & 24.9  & 18.2 & 17.8 & \underline{20.5} & 23.7 & 14.3 & \underline{4.4} & \textbf{22.4} & \underline{13.6} & \underline{19.5} \\
        \hline
        gte-Qwen2-7B-instruct & 100 & 26.1 & 30.4 & 16.8 & 26.6 & 18.7 & 18.5 & 16.5 & 18.7 & \textbf{20.8} & 2.9 & 20.4 & 12.4 & 19.1 \\
        \hline
        \rowcolor{orange!30}
        \multicolumn{15}{c}{\projectname{} models (MATH+NuminaMath, all query types)} \\ \hline
        gte-Qwen2-7B-instruct & 100 & 25.5 & \underline{31.8} & \textbf{19.3} & \textbf{28.8} & \underline{22.0} & \underline{19.8} & 20.1 & 17.1 & 11.9 & 1.6 & 18.9 & \textbf{14.3} & 19.3 \\
        \hline
        Qwen2.5-7B-instruct & 100 & 25.0 & 31.1 & \underline{17.3} & 26.4 & 21.1 & 19.5 & \textbf{21.3} & 21.5 & \underline{16.1} & \textbf{5.1} & \underline{21.7} & \textbf{14.3} & \textbf{20.0}\\
         \hline
    \end{tabular}
    \end{adjustbox}
   
    \caption{nDCG@10 performance of different rerankers on \textsc{Bright}. Reranking is performed on the top-10 or top-100  results retrieved using BM25 with the question as the query. Results with $^\star$ are  from~\cite{su2024brightrealisticchallengingbenchmark}.}
    \label{table:rerankbm25}
\end{table*}

%% file: Tables/table-reranking-gpt4o.tex
\begin{table*}[!th]
\scriptsize
\centering
\begin{adjustbox}{max width=\textwidth}
\begin{tabular}{l|ccccccc|cc|ccc|c}
\hline
 & \multicolumn{7}{c|}{StackExchange} & \multicolumn{2}{c|}{Coding} & \multicolumn{3}{c|}{Theorem-based} & \multirow{2}{*}{\textbf{Avg.}} \\
 \cline{2-8} \cline{9-10} \cline{11-13}
 & Bio. & Earth. & Econ. & Psy. & Rob. & Stack. & Sus. & Leet. & Pony & AoPS & TheoQ. & TheoT. & \\
\hline
BM25$^\star$ & 19.2 & 27.1 & 14.9 & 12.5 & 13.5 & 16.5 & 15.2 & \textbf{24.4} & 7.9 & 6.0 & 13.0 & 6.9 & 14.8 \\
BM25 on GPT-4o CoT & \textbf{53.6} & \textbf{53.6} & 24.3 & 38.6 & 18.8 & 22.7 & 25.9 & \underline{19.3} & 17.7 & 3.9 & 18.9 & 20.2 & 26.5 \\
\hline
\rowcolor{brown!20}
\multicolumn{14}{c}{Reranking on GPT-4o CoT k=100} \\
\hline
MonoT5-3B$^\star$ & 16.0 & 24.0 & 17.7 & 19.5 & 8.0 & 10.5 & 19.5 & 17.2 & 29.2 & \textbf{7.1} & 20.3 & 12.0 & 16.8 \\
RankLLaMA-7B$^\star$ & 17.5 & 15.5 & 13.1 & 13.6 & 17.9 & 6.9 & 16.9 & 8.4 & \textbf{46.8} & 2.2 & 4.5 & 3.5 & 13.9 \\
Rank1-7B~\cite{weller2025rank1testtimecomputereranking} & \underline{48.8} & \underline{36.7} & 20.8 & 35.0 & 22.0 & 18.7 & \textbf{36.2} & 12.7 & 31.2 & \underline{6.3} & \textbf{23.7} & \underline{37.8} & \textbf{27.5} \\
\hline
\rowcolor{orange!30}
 \multicolumn{14}{c}{\projectname{} Models Reranking on GPT-4o CoT results k=100 (MATH, all query types)} \\
 \hline
  Qwen2.5-7B-instruct  & 40.8 & 31.8 & \underline{25.0} &  39.7 & 21.6 & 25.7 & 27.2 & 17.3 & 29.9 & 1.6 & \underline{22.4} & 36.9 & \underline{26.7} \\
gte-Qwen2-7B-instruct  & 36.0 & 29.3 & 23.2 & 40.0 & 23.2 & 24.1 & 22.2 & 17.8 & \underline{34.9} & 1.5 & 20.4  & 35.3 & 25.7 \\
\hline
\rowcolor{orange!30}
\multicolumn{14}{c}{\projectname{} Models Reranking on GPT-4o CoT results k=100 (MATH+NuminaMath, all query types)} \\
\hline
Qwen2.5-7B-instruct & 37.0 & 32.4 & \textbf{25.5} & \underline{41.5} & \underline{24.9} & \textbf{26.7} & \underline{28.1} & 12.2 & 28.8 &  2.7 & 21.7 & \textbf{39.1} & \underline{26.7} \\
gte-Qwen2-7B-instruct  & 36.9 & 31.2 & 24.9 & \textbf{43.1} &  \textbf{26.4} & \underline{26.3} & 26.1 & 16.6 & 26.6 & 0.4 & 18.9 & 36.1 & 26.1 \\
\hline
\end{tabular}
\end{adjustbox}
\caption{nDCG@10 performance of rerankers  on \textsc{Bright} using questions as retrieval queries. 
 The first-stage results are obtained by BM25 using GPT-4o CoT reasoning as queries. Results with $^\star$ are taken from~\cite{su2024brightrealisticchallengingbenchmark}. }
\label{tab:rerankperformance}
\end{table*}

%% file: Tables/table-e2e.tex
\begin{table}[!ht]
\centering
\resizebox{!}{0.21\linewidth}{%
\begin{tabular}{lc}
\midrule
\textbf{Method} & \textbf{Accuracy(\%)} \\
\midrule
Base model (no retrieval) & 71.0 \\ \hline
In-context RAG with \projectname{}  & 75.0 \\
In-context RAG with RepLLama & 72.6 \\
In-context RAG with gold theorems & 77.6 \\ \hline
MCTS with only OST action & 75.0 \\
MCTS with OST + RepLLaMA retrieval & 78.9 \\
MCTS with OST + \projectname{} retrieval &  80.2 \\ 
MCTS with OST + gold theorems & 81.5 \\
\bottomrule
\end{tabular}
}
\caption{QA performance of Qwen-2.5-7B-Instruct on the TheoremQA theorems split of  \textsc{Bright} in different settings of retrieval augmentation.}
\label{tab:accuracy_comparison}
\end{table}

%% file: conclusion.tex
\section{Conclusion}
We introduce \projectname, a suite of retrievers and rerankers designed for reasoning-intensive relevance ranking. Our approach employs a retrieval-augmented reasoning framework based on Monte Carlo Tree Search (MCTS) to generate high-quality training data. \projectname achieves substantial improvements in both retrieval and reranking performance over state-of-the-art models on reasoning-intensive benchmarks,   demonstrating strong generalizability and significantly higher data efficiency compared to existing methods.

%% file: limitation.tex
\section{Limitations}

While \projectname achieves strong performance on reasoning-intensive retrieval tasks, it has a few limitations.
 First, our training approach primarily focuses on examples where the retriever reasons over a single document in isolation. A promising direction for future work is to develop retrievers capable of reasoning over multiple documents jointly, where the relevance of each document is informed by the content of others, similar to the requirements in multi-hop QA tasks.

 Second, \projectname models focus on producing relevance scores without generating explicit explanations for document retrieval. Future work could explore the development of reasoning-aware retrievers that offer greater transparency and interpretability by generating explanations for their retrieval decisions, all while maintaining their efficiency for the first-stage ranking.

 Lastly, we use rewards in our MCTS framework, based on the final answer of the math reasoning datasets. However, incorrect CoT  reasoning path for solving a mathematical question can lead to a correct final answer, thus our training data can be noisy.

%% file: Appendix_A.tex
\input{detailed-related-work}

\section{Sampling MCTS Rollout Solutions}
\label{sec:MCTSrollouts}
The MCTS proceeds with multiple iterations of four main processes: \textit{selection}, \textit{expansion}, \textit{simulation} and \textit{backpropagation}.
To balance exploration and the exploitation, the \textit{selection} step starts from the root node and uses Upper Confidence Bounds applied to trees (UCT) to traverse through child nodes, continuing until a leaf node is reached. Formally, we select the node with maximum UCT value at each branch of the traversal: \[\mathrm{UCT}(s,a) = \frac{Q(s,a)}{N(s,a)} + c\sqrt{\frac{ln (N_{\mathrm{parent}}(s))}{N(s,a)}} \] where $N(s,a)$ is the number of times node $s$ has been visited till now and $Q(s,a)$ is the expected reward of node $s$ under action $a$ and $c$ is a hyperparameter. If the leaf node is not a \textit{terminal} node, the \textit{expansion} step adds child nodes to the leaf node to represent potential future actions. The \textit{simulation} step selects one of the newly added child nodes at random and performs rollouts/simulations by selecting actions randomly until we reach a terminal node $t$. Based on whether the terminal node $t$ reaches the correct gold answer $G$, we calculate a reward value $R(t)$ and update $Q(s,a)$ values for all the nodes $s_i$ in the collected solution trajectory $M \oplus s_1 \oplus s_2 \oplus \cdots \oplus s_{t}$ as: \[Q(s_i,a) = Q(s_i,a) + R(t)\] The $N(s,a)$ values are also incremented as \[N(s_i,a) = N(s_i,a) + 1\].

\section{Math Datasets for Data Generation}
\label{details:mathdataset}

To construct our training dataset, we leverage mathematical reasoning benchmarks consisting of natural language math questions paired with a correct answer, typically represented as a boxed numerical value in \LaTeX. Our data generation pipeline does not require access to gold step-by-step solutions.

\paragraph{MATH} The MATH dataset comprises mathematical problems across 8 different subject types (Prealgebra, Precalculus, Algebra, Geometry, Intermediate Algebra, Counting and Probability, and Number Theory) and five difficulty levels (from 1 to 5, where 1 denotes the easiest).
\paragraph{NuminaMath.}NuminaMath is a large-scale collection comprising 860K competition-level math problems paired with solutions.  Our MCTS leverages the {OrcaMATH}, {AMC}, {AIME}, {Chinese K-12 Exam}, {Olympiad}, and {AOP forum} splits from {NuminaMath}.

\input{Tables/table-MCTS-params}
\section{MCTS Parameters}
\label{mctsparams}
Table~\ref{table:MCTS-params} shows the values of hyperparameters for the MCTS algorithm.

\input{Tables/table-statistics-training-data}
\section{Statistics of Synthesized Data}
\label{sec:stats-training-data}
Table~\ref{table:statistics} shows the number of synthesized samples from each dataset for mathematical problem solving. 

\section{Retriever Training Details}
\label{sec:appendix_a_trainingloss}
We append an end-of-sequence token (\texttt{EOS} token) to the input query or document to form the input sequence to our base LLM. Thus, the vector embedding of a query or a document (denoted as $t$) is computed as:

\[
{E}_{t} = \mathrm{Decoder}({t}_1 {t}_2\ \cdots\ {t}_k \texttt{<eos>})[-1],
\]
where $\mathrm{Decoder}$($\cdot$) represents the LLM model (such as Qwen or Llama), which returns the last layer token representations for each input token. We take the representation of the end-of-sequence token as the representation of the input sequence $t_1, \ldots, t_k$, which can be either a query $q$ or a document $d$. Relevance of $d$ to $q$ is computed using the cosine similarity of their corresponding dense representations $E_q$ and $E_d$ as:

\[
s( q, d) = \cos(E_q, E_d). 
\]
The model is then optimized end-to-end using the InfoNCE loss:
\begingroup
\scriptsize
\begin{equation}
 \mathcal{L}(q, p, D^-) = -\log \frac{\exp(s(q,p))}{\exp(s(q,p)) + \sum_{d^- \in D^-} \exp(s(q,d^-))},   
\end{equation}
\endgroup
where $p$ denotes a document relevant to the query $q$ (based on human annotations), while $D^-$ is the set of negative (non-relevant) documents.

\section{Reranker Training Details}
\label{sec:appendix_a_trainingloss_reranker}
Our reranker model is trained as pointwise reranker. The input to the model is a concatenation of the query and a candidate document, with the model generating a score that indicates the relevance of the document to the query \cite{nogueira-etal-2020-document}. 

In more detail, our model reranks a query-document pair as shown below:\\ \\
\indent $\mathrm{input}  = \text{query:} \: \{q \}\; \text{document:} \: \{d\} $ \texttt{<eos>}\\
\indent $s(q, d) = \mathrm{Linear}(\mathrm{Decoder}(\mathrm{input})[-1])$ \\

Here, $\mathrm{Decoder}(\cdot)$ represents the LLM model (such as Qwen or Llama), which returns the last layer token representations for each input token , and $\mathrm{Linear}(\cdot)$ is a linear projection layer that maps the final hidden state corresponding to the end-of-sequence token to a scalar relevance score. The training uses same loss, used for retriever training, with no use of in-batch negatives.

\input{Tables/table-hyper-params-ranking}

\section{Retriever/Reranker Training Hyperparameters}
\label{sec:appendix_a_hyperparams}
We used \textit{Tevatron}~\cite{ma2025tevatron} package for training. The hyperparameters used for finetuning retrieval and reranking models are presented in Table~\ref{table:ranking-hyper-params}.


\input{Tables/table-bright-q-prec}
\input{Tables/table-bright-q-rec}

\section{Retrieval Performance on \textsc{Bright}}
\label{prec-rec-on-bright}

In addition to nDCG performance of retrieval models, we compare our \projectname{} models with baselines in terms of precision at top 10 documents in Table~\ref{table:retrieval-bright-CoT-queries_precision10} and recall at top 10 documents in Table~\ref{table:retrieval-bright-CoT-queries_recall10} when questions are used as retrieval queries.

\input{ablation}

\input{analysis}

%% file: detailed-related-work.tex
\section{Detailed Discussion of Related Works}

\subsection{Mathematical Information Retrieval} 

Math Information Retrieval (Math IR) has been extensively studied within the IR community, focusing on the task of retrieving relevant mathematical documents such as theorems, formulas, similar questions, or textbooks to solve a given math problem. Early models, such as Approach0~\cite{zhong2021approach}, perform retrieval by using structural similarities between the formulas in  queries and  documents.

MathBERT~\cite{zhong-etal-2022-evaluating} pre-trained a cross-encoder BERT-base model  on a corpus of 1.69 million math documents containing both text and formulas. Several approaches~\cite{zhong-etal-2022-evaluating}  explored hybrid methods combining dense neural retrievers with structural and lexical retrievers to improve retrieval performance.

The NTCIR-10 Math Pilot Task \cite{Aizawa2013NTCIR10MP} marked one of the first collaborative efforts to establish evaluation frameworks for \textit{mathematical formula search}, while the \textsc{ARQMath} Lab tasks \cite{Mansouri2022OverviewOA} extended \textsc{Math IR} to a Community Question Answering (CQA) setting, utilizing user-generated data from \textit{Math StackExchange}.

More recently, datasets such as \textbf{\textsc{Bright}}~\cite{su2024brightrealisticchallengingbenchmark} and RAR-B~\cite{xiao2024rarbreasoningretrievalbenchmark} have been introduced as evaluation benchmarks for Math IR, focusing on tasks including \textit{Theorem Retrieval}, \textit{Similar Questions Retrieval}, and \textit{Answer Retrieval}.
In this work, we adopt both \textbf{\textsc{Bright}} and RAR-B as evaluation datasets to assess the effectiveness of our trained retriever.

\subsection{CoT Reasoning}
\label{subsec:cotreasoning}
IRCoT~\cite{trivedi2023interleavingretrievalchainofthoughtreasoning} uses the last reasoning step as a retrieval query, conditioning future steps on retrieved documents. ITER-RETGEN~\cite{shao-etal-2023-enhancing} alternates between retrieval and generation, showing improvements in tasks such as multi-hop QA and fact verification. REACT~\cite{yao2023react} iteratively generates (thought, action, observation) sequences, using intermediate reasoning to drive retrieval. Toolformer~\cite{schick2023toolformer} employs self-supervised training to help models autonomously determine when to invoke external retrieval tools, like the \textit{Wikipedia Search API}.

\subsection{Reasoning-Based Re-Ranking Models}
\label{subsec:reasoningrerank}
\textsc{Rank-1}\cite{weller2025rank1testtimecomputereranking} employs knowledge distillation from DeepSeek-R1, extracting over 600K reasoning examples from the \textsc{MS-MARCO} dataset\cite{nguyen2017ms} to train a re-ranking model. Similarly, InteRank~\cite{samarinas2025distillationrefinementreasoningsmall} uses reinforcement learning to train a 3B-parameter re-ranking model, generating reasoning explanations alongside relevance scores for (query, document) pairs. Despite their improved retrieval quality, these re-ranking models remain inherently dependent on the first-stage retriever’s candidate set, typically derived from lexical or semantic matching, thus limiting their effectiveness on reasoning-intensive retrieval tasks. In contrast, \projectname develops a dedicated first-stage retriever from pretrained language models, including \textit{Qwen2.5}~\cite{yang2024qwen2}, \textit{Llama 3.1}~\cite{grattafiori2024llama}, and \textit{gte-Qwen2-7B-instruct}~\cite{li2023towards}, tailored explicitly to reasoning-intensive search scenarios.

%% file: Tables/table-MCTS-params.tex
\begin{table}[!ht]
\centering
\label{mcts-Hyper_params_table}
\begin{tabular}{l|l}
\hline
\textbf{Parameter} & \textbf{Value} \\
\hline
No of RT nodes per action (Top-$k$)& 5\\
No of OST nodes added (per action) &2\\
No of QG nodes added (per action) &1\\
No of rollouts & 16\\
Max depth &6\\
MCTS Exploration weight $C$ & 2\\
MCTS Weight Scheduler & const \\
LLM$_\mathrm{gen}$ temperature & 0.8\\
LLM$_\mathrm{gen}\hspace{2mm}$top-$k$ & 40\\
LLM$_\mathrm{gen}\hspace{2mm}$top-$p$ & 0.95\\
BF16 & Enabled \\
GPUs & 2 A100s\\
\hline
\end{tabular}
\caption{MCTS Parameters}
\label{table:MCTS-params}
\end{table}

%% file: Tables/table-statistics-training-data.tex
\begin{table}[!ht]
\centering
\resizebox{\linewidth}{!}{
\begin{tabular}{l|l}
\midrule
\textbf{Dataset (query type)} & \textbf{\# of Samples} \\
\midrule
MATH ($q_\mathrm{llmq}$)& 18,586\\
MATH ($q_\mathrm{CoT}$)& 7,312 \\
MATH ($q_\mathrm{question}$)& 7,312 \\
MATH ($q_\mathrm{lexical}$)& 9,910 \\
\midrule
MATH (all queries) & 43,120 \\
\midrule
NuminaMATH ($q_\mathrm{llmq}$)& 39,639\\
NuminaMATH ($q_\mathrm{CoT}$)& 24,280\\
NuminaMATH ($q_\mathrm{question}$)& 10,241\\
NuminaMATH ($q_\mathrm{lexical}$)& 4,158 \\
\midrule
NuminaMATH (all queries) & 78,318 \\
\midrule
MATH+NuminaMATH (all queries) & 121,438 \\
\bottomrule
\end{tabular}
}
\caption{Statistics of synthesized data for retrieval training.}
\label{table:statistics}
\end{table}

%% file: Tables/table-hyper-params-ranking.tex
\begin{table}[!ht]
\centering
\label{Hyper_params_table}
\begin{tabular}{l|l}
\hline
\textbf{Parameter} & \textbf{Value} \\
\hline
Train Group Size& 12\\

 Warmup Steps&28\\
Per Device Train Batch Size & 2\\
Gradient Accumulation Steps & 16\\
 DDP Timeout&1800\\
Temperature& 0.01\\
Learning Rate& 1e-4\\
LR Scheduler Type & Linear\\
Number of Train Epochs & 1\\
BF16 & Enabled \\
GPUs & 2 A100s\\
\hline
\end{tabular}
\caption{Retriever Training Hyperparameters.}
\label{table:ranking-hyper-params}
\end{table}

%% file: Tables/table-bright-q-prec.tex
\begin{table*}[ht]
\centering
\scriptsize
\resizebox{\textwidth}{!}{
\begin{tabular}{l|ccccccc|cc|ccc|c}
\hline
& \multicolumn{7}{c|}{StackExchange} & \multicolumn{2}{c|}{Coding} & \multicolumn{3}{c|}{Theorem-based} & \multirow{2}{*}{\textbf{Avg.}} \\
\cline{2-8} \cline{9-10} \cline{11-13}
& Bio & Earth & Econ & Psy & Rob & Stack & Sus & Leet & Pony & AoPS & TheoQ & TheoT & \\
\hline
\rowcolor{brown!20}
\multicolumn{14}{c}{Sparse and Open-source Baselines} \\
\hline
BM25$^\star$ & 7.6 & 12.4 & 7.1 & 6.0 & 5.6 & 8.0& 6.1 & 6.0& 7.9 &3.1 & 2.2& 1.3&6.1 \\
Qwen$^\star$  & 13.5 &  14.1& 8.2 & 11.2& 5.8 & 10.1 & 6.1& 6.3 &  & 9.7 & 7.1 & 6.2 & 7.3 \\
GritLM$^\star$  & 11.1 & 12.7 & 9.2 & 10.8 & 6.8 & 6.0 & 6.8& 7.5 & 17.9 &4.6  & 5.7 & 5.3 & 8.7 \\
Inst-XL$^\star$  &10.0 & 13.8& 10.6 & 11.0 & 6.7 &  8.8 & 9.2&6.3 & 4.6 & 4.7 & 3.3 & 1.9 & 7.6\\
E5$^\star$  & 8.9 &10.2 & 8.1 &8.6 & 7.2 & 4.6 & 6.9 & 6.9 &4.9 &4.2 &5.7  &6.5 &6.9  \\
\hline
\rowcolor{red!20}
\multicolumn{14}{c}{Proprietary Baselines} \\
\hline
Google$^\star$  & 10.3 &12.2  &8.9 &11.4  &5.6 &8.3  &7.9   &6.9  &3.6 &5.0 &4.8  &4.2  &7.4  \\
Voyage AI$^\star$  & 11.0&9.9 &9.6 &11.0 &5.6  &7.3  &7.5  & 7.5 &1.1  &5.0 &5.6  &3.2  & 7.0  \\
\hline
\rowcolor{orange!30}
\multicolumn{14}{c}{\projectname{} Models (MATH dataset, $q_\mathrm{llmq} + q_\mathrm{CoT} + q_\mathrm{lexical}$)} \\
\hline
 Qwen2.5-7B-instruct  & 11.6 &11.7  &6.8  & 9.9& 5.3 & 7.9 &6.7  &8.5  & 8.0 & 8.2 & 10.2 & 9.9 &\textbf{8.7}\\
\hline
\rowcolor{orange!30}
\multicolumn{14}{c}{\projectname{} Models (MATH$+$NuminaMath datasets, all query types)} \\ \hline
 Qwen2.5-7B-instruct  & 12.2&  11.5& 7.8& 10.6&5.7 &7.4 & 6.8& 8.1& 6.2& 6.0& 9.2 & 11.3& 8.6 \\
\hline
\end{tabular}
}
\caption{Precision@10 performance using question as queries for retrieval over \textsc{Bright}.}
\label{table:retrieval-bright-CoT-queries_precision10}
\end{table*}

%% file: Tables/table-bright-q-rec.tex
\begin{table*}[ht]
\centering
\scriptsize
\resizebox{\textwidth}{!}{
\begin{tabular}{l|ccccccc|cc|ccc|c}
\hline
& \multicolumn{7}{c|}{StackExchange} & \multicolumn{2}{c|}{Coding} & \multicolumn{3}{c|}{Theorem-based} & \multirow{2}{*}{\textbf{Avg.}} \\
\cline{2-8} \cline{9-10} \cline{11-13}
& Bio & Earth & Econ & Psy & Rob & Stack & Sus & Leet & Pony & AoPS & TheoQ & TheoT & \\
\hline
\rowcolor{brown!20}
\multicolumn{14}{c}{Sparse and Open-source Baselines} \\
\hline
BM25$^\star$ & 21.8 & 31.4 & 16.8 & 15.5 & 19.4 & 16.8 & 21.1 & 29.5 & 3.6 & 6.0 & 11.4 & 9.0 & 16.9 \\
Qwen$^\star$  & 38.2 & 40.6 & 18.5 & 29.5 & 14.5 & 22.4 & 17.4 & 32.1 & 4.6 & 14.8 & 30.0 & 39.4 & 25.2 \\
GritLM$^\star$  & 30.3 & 38.8 & 18.3 & 26.9 & 21.3 & 15.1 & 23.4 & 36.3 & 8.2 & 9.4 & 26.2 & 26.6 & 23.4 \\
Inst-XL$^\star$  & 27.3 & 38& 25.4 & 35.6 & 22 & 21.1 & 23.9 & 31.8 & 2.5 & 8.9 & 16.6 & 9.8 & 21.9 \\
E5$^\star$  & 22.0 & 29.4 & 18.4 & 18.3 & 18.7 & 11.9 & 23.0 & 34.6 & 2.4 & 8.2 & 27.2 & 34.8& 20.7 \\
\hline
\rowcolor{red!20}
\multicolumn{14}{c}{Proprietary Baselines} \\
\hline
Google$^\star$  & 26.1 & 36.9 & 20.6 & 31.4 & 17.7 & 21.6 & 23.7  & 33.5 & 1.9 & 10.4 & 24.0 & 22.1 & 22.5 \\
Voyage AI$^\star$  & 29.3 & 31.2 & 21.0 & 31.0 & 15.0 & 17.5 & 20.5 & 41.5 & 0.6 & 8.7 & 28.5 & 15.4 & 21.7 \\
\hline
\rowcolor{orange!30}
\multicolumn{14}{c}{\projectname{} Models (MATH dataset, $q_\mathrm{llmq} + q_\mathrm{CoT} + q_\mathrm{lexical}$)} \\
\hline
 Qwen2.5-7B-instruct  & 30.6 & 36.9 & 19.5 & 30.1 & 18.5 & 28.6 & 21.1 & 46.5 & 6.4 & 18.1 & 48.3 & 56.0 & 30.1\\
 gte-Qwen2-7B  & 41.6 & 42.5 & 26.1 & 41.2& 18.6 &32.5& 31.7& 44.7 & 3.4 & 13.4& 32.5& 47.9& \textbf{31.3}\\
 Llama3.1-8B-Instruct  & 35.7 & 33.5 & 18.4  & 36.4 & 14.8 & 28.5 & 20.3& 45.2& 6.4 & 7.7 & 37.0 & 43.4 & 27.3 \\
\hline
\rowcolor{orange!30}
\multicolumn{14}{c}{\projectname{} Models (MATH$+$NuminaMath datasets, all query types)} \\ \hline
 Qwen2.5-7B-instruct  & 31.3 & 34.2 & 21.5 & 31.9 & 18.8 & 30.5 & 21.5 & 45.6 & 5.1 & 11.7 & 44.7 & 63.8& 30.1 \\
\hline
\end{tabular}
}
\caption{Recall@10 performance using question as queries for retrieval over \textsc{Bright}.}
\label{table:retrieval-bright-CoT-queries_recall10}
\end{table*}

%% file: ablation.tex

 \noindent \input{Tables/table-ablation-data}
 \noindent \input{Tables/table-ablation-llm-size}

\section{Ablation Studies}
\label{ablations_models}
\textbf{Effect of query types in synthesized samples.} 
We evaluate the impact of different query types in the  synthesized retrieval data by comparing variants of our \projectname{} models trained on subsets including specific query types. 
Table~\ref{table:queryablations} summarizes the results of this ablation on \textsc{Bright}.

We observe a consistent improvement in retrieval performance as additional query types are included during training. These results highlight the \emph{complementary} nature of the diverse query types in our synthesized training data.

 \noindent \textbf{Impact of base-LLM size.} 
Table~\ref{table:sizeablations} presents the performance of our \projectname{} retrievers using Qwen-2.5-instruct models of varying sizes (3B, 7B, 14B). We observe consistent gains with model scaling, with the 14B model outperforming the 7B variant by 2 points on average.

%% file: Tables/table-ablation-data.tex
\begin{table*}[ht]
\centering
\scriptsize
\resizebox{\textwidth}{!}
{\begin{tabular}{l|ccccccc|cc|ccc|c}
\hline
\multirow{2}{*}{Data of \projectname{}- } & \multicolumn{7}{c|}{StackExchange} & \multicolumn{2}{c|}{Coding} & \multicolumn{3}{c|}{Theorem-based} & \multirow{2}{*}{\textbf{Avg.}} \\
\cline{2-2} \cline{2-8} \cline{9-10} \cline{11-13}
 & Bio & Earth & Econ & Psy & Rob & Stack & Sus & Leet & Pony & AoPS & TheoQ & TheoT & \\
\hline
$q_\mathrm{llmq}$ & \textbf{25.8} & \textbf{32.0} & 15.8 & 23.3 & 14.8 & 18.7 & 15.7 & 29.9 & \textbf{16.4} & 13.0 & 38.8 & 35.4 & 23.3 \\
$q_\mathrm{llmq} + q_\mathrm{CoT}$ & 24.5 & 26.9 & \textbf{17.0} & \textbf{25.6} & \textbf{14.8} & 21.1 & \textbf{17.1} & 30.6 & 9.9 & 12.8 & 42.5 & 38.5 & 23.4 \\
 $q_\mathrm{llmq} + q_\mathrm{CoT} + q_\mathrm{lexical}$ & 25.4 & 30.0 & 16.7 & 25.3 & 14.0 & \textbf{21.3} & 16.3 & \textbf{37.0} & 8.2 & \textbf{15.7} & \textbf{42.7} & \textbf{44.4} & \textbf{24.6}  \\

\hline
\end{tabular}
}
\caption{nDCG@10 performance of \projectname{} on \textsc{Bright} when retrievers are trained using different types of samples generated from MATH.
The original math question is used as the query. \projectname{} Qwen2.5-7B-instruct models are used for the ablations.}
\label{table:queryablations}
\end{table*}

%% file: Tables/table-ablation-llm-size.tex
\begin{table*}[ht]
\scriptsize
\centering
\resizebox{\textwidth}{!}
{\begin{tabular}{l|ccccccc|cc|ccc|c}
\hline
\projectname{} Models & \multicolumn{7}{c|}{StackExchange} & \multicolumn{2}{c|}{Coding} & \multicolumn{3}{c|}{Theorem-based} & \multirow{2}{*}{\textbf{Avg.}} \\
\cline{2-2} \cline{2-8} \cline{9-10} \cline{11-13}
(all query types) & Bio & Earth & Econ & Psy & Rob & Stack & Sus & Leet & Pony & AoPS & TheoQ & TheoT & \\
\hline
Qwen2.5-3B-instruct  & 26.9 & 29.7 &16.5 & 24.8 & 13.3& 19.3&  15.2&  37.5 &\textbf{10.6}  &9.4 & 34.9 & 36.9 & 22.9\\
 Qwen2.5-7B-instruct  & 25.1 & 28.3 & \textbf{18.2} & 25.9 & 15.3 & 22.2 & 16.3 & 35.9 & 6.9 & 10.4 & \underline{40.8} & \textbf{47.1} & \underline{24.4} \\
Qwen2.5-14B-instruct & \textbf{30.9} & \textbf{31.5} & 17.3 &  \textbf{27.9} & \textbf{16.0} & \textbf{24.2} & \textbf{16.4} & \textbf{40.9} & 10.2 & \textbf{12.7} & \textbf{42.6} & 45.8 & \textbf{26.4} \\
\hline
\end{tabular}
}

\caption{nDCG@10 performance of \projectname{} on \textsc{Bright} when retrievers are trained using different sizes of Qwen-2.5-instruct models (3B, 7B, and 14B). The original math question is used as the query.}
\label{table:sizeablations}
\end{table*}

%% file: analysis.tex
\section{Analysis}
\label{analysis:appendix}
\input{query_types}
We present examples from our synthesized training data, along with qualitative examples of \projectname{} from the \textsc{Bright} evaluation. 
We highlight representative cases where \projectname{} successfully retrieves the correct theorem, as well as illustrative failure cases to analyze its limitations.

\paragraph{Examples of queries from Retrieval Training Dataset.} 
We present a qualitative example of different query types ($q_\mathrm{llmq}, q_\mathrm{CoT}$ and $q_\mathrm{lexical}$) for a question $M$ and positive theorem $p$ from our dataset in Figure~\ref{figure:exampledataset}. As illustrated in the figure, $q_\mathrm{CoT}$ captures the broader reasoning context of the mathematical question, thereby encouraging the model to learn a more challenging retrieval task. In contrast, $q_\mathrm{lexical}$ exhibits high lexical overlap with the associated positive document, which helps for queries which do not require reasoning. The query $q_\mathrm{llmq}$ strikes a balance between these two extremes, as it is generated using both $M$ and $P$, effectively combining elements of both contextual reasoning and lexical similarity.
\input{successcases_example}
\paragraph{Examples from \textsc{Bright} evaluation.}
We present examples from TheoremQA theorems split of BRIGHT, to qualitatively compare \projectname{} with other baseline retrievers. For the example of Figure~\ref{qualanalysis}, \projectname{} successfully performs nuanced reasoning to successfully retrieve the gold theorem, whereas a strong baseline, Qwen2, fails to do so. However, challenges still remain.
\input{failure_cases}Figure~\ref{qualanalysis_failure} presents an example, where the question is about directionality in a family tree graph. Both retrievers Qwen2 and \projectname{}, fail to capture the query's focus on \textit{acyclicity} and instead match on the topic of \textit{structural hierarchy} in the graph, thus leading to an incorrect retrieval. This highlights the need for further advancements in reasoning-aware retrieval methods.
\paragraph{Analysis of Coding example from Leetcode}\label{analysis_code}In Figure~\ref{codingexample}, we present a qualitative example from Leetcode, which shows how recognizing mathematical subtructure in a coding problem can help \projectname{} retrievers. Retrieving relevant documents for the coding task, requires application of mathematical reasoning which \projectname{} models excel in, whereas BM25 only performs lexical matching based on the word \textit{rectangle}, which results in an incorrect retrieval. 
\clearpage
\input{codingexample}

%% file: query_types.tex
\begin{figure*}[ht]
\begin{tcolorbox}[
    enhanced,
    sharp corners,
    colback=white,
    colframe=black,
    boxrule=0.8pt,
    width=\textwidth,
    breakable,
    left=0pt,
    right=0pt,
    top=0pt,
    bottom=0pt,
    boxsep=0pt,
    arc=0mm,
]
\begin{tcolorbox}[
    colback=blue!10,
    colframe=blue!60,
    width=\textwidth,
    boxrule=0pt,
    arc=0mm,
    left=6pt,
    right=6pt,
    top=6pt,
    bottom=6pt,
    title=\textbf{\large Example},
    coltitle=black,
    fonttitle=\bfseries,
]
\textbf{Question ($M$)}: Let \( f(x) \), \( g(x) \), and \( h(x) \) be linear functions. Define
\[
j(x) = \max\{f(x), g(x), h(x)\}, \qquad k(x) = \min\{f(x), g(x), h(x)\}.
\]
Suppose the graph of \( y = j(x) \) over \( -3.5 \le x \le 3.5 \) consists of three connected line segments: it increases linearly to \( (-2, 2) \), remains constant at \( y = 2 \) from \( x = -2 \) to \( x = 2 \), and then increases linearly from \( (2, 2) \) to \( x = 3.5 \). Let \( \ell \) denote the total length of the graph of \( y = k(x) \) over the same interval. Find \( \ell^2 \). \\ \\
\textbf{Positive document} ($p$): Linear Real Function - A real function \( f \) is called linear if it has the form \( f(x) = ax + b \), where \( a \) and \( b \) are real numbers.
\end{tcolorbox}
\begin{tcolorbox}[
    colback=red!10,
    colframe=red!60,
    width=\textwidth,
    boxrule=0pt,
    arc=0mm,
    left=6pt,
    right=6pt,
    top=6pt,
    bottom=6pt,
    title=\textbf{\large Different Query Types},
    coltitle=black,
    fonttitle=\bfseries,
]
\textbf{LLM generated query} ($q_\mathrm{llmq}$): Calculate the length of the graph of a piecewise linear function defined by the minimum of three linear functions over a given interval. \\ \\
\textbf{CoT reasoning query ($q_\mathrm{CoT}$)}: {Question} +  To determine the length of the graph of \( y = k(x) \) for \( -3.5 \leq x \leq 3.5 \), we first need to understand the behavior of \( j(x) \), the maximum of the three linear functions \( f(x) \), \( g(x) \), and \( h(x) \). From the graph, we can see that \( j(x) \) transitions between the three linear functions at points where two of the functions are equal. The graph of \( j(x) \) has horizontal segments and linear segments connecting the points where the maximum changes. \\ \\ 
\textbf{Lexical query ($q_\mathrm{lexical}$)}: What is the definition of a linear real function and what form must it take for all real numbers $x$? 
\end{tcolorbox}
\end{tcolorbox}

\caption{Examples of different query types from our retrieval training dataset built for the given math question.}
\label{figure:exampledataset}
\end{figure*}

%% file: successcases_example.tex
\begin{figure*}[ht]
\begin{tcolorbox}[
    enhanced,
    sharp corners,
    colback=white,
    colframe=black,
    boxrule=0.8pt,
    width=\textwidth,
    breakable,
    left=0pt,
    right=0pt,
    top=0pt,
    bottom=0pt,
    boxsep=0pt,
    arc=0mm,
]
\begin{tcolorbox}[
    colback=blue!10,
    colframe=blue!60,
    width=\textwidth,
    boxrule=0pt,
    arc=0mm,
    left=6pt,
    right=6pt,
    top=6pt,
    bottom=6pt,
    title=\textbf{\large Example},
    coltitle=black,
    fonttitle=\bfseries,
]
\textbf{Query:} Everyone that you invite to a party will be either a fan of \textit{football} or a fan of \textit{basketball}, but never both. What is the smallest number of guests you need to invite to ensure that there are either \textbf{3 people} who are all fans of football or \textbf{3 people} who are all fans of basketball?\\ \\
\noindent\textbf{Gold Theorem Id}: 7627\\ 
\textbf{Gold Theorem}: \textit{Ramsey Theorem: Ramsey's Theorem guarantees that for any edge-coloring of a sufficiently large complete graph, there exists a monochromatic complete subgraph. More formally, for integers $n_1, n_2, \ldots, n_c$, there exists a Ramsey number $R(n_1, \ldots, n_c)$ such that any $c$-coloring of a complete graph on $R(n_1, \ldots, n_c)$ vertices contains a monochromatic $K_{n_i}$ in some color $i$.}
\textit{(Note that the gold theorem has no lexical overlap with the original question.)}
\end{tcolorbox}
\begin{tcolorbox}[
    colback=red!10,
    colframe=red!60,
    width=\textwidth,
    boxrule=0pt,
    arc=0mm,
    left=6pt,
    right=6pt,
    top=6pt,
    bottom=6pt,
    title=\textbf{\large Qwen2 retriever},
    coltitle=black,
    fonttitle=\bfseries,
]
\textbf{Top Retrieved Theorem}: \textit{Pigeonhole Principle: Let $S$ be a finite set with $n$ elements, partitioned into $k$ subsets $S_1, S_2, \ldots, S_k$. Then at least one subset $S_i$ satisfies:}
\[
|S_i| \ge \left\lceil \frac{n}{k} \right\rceil
\]
\textcolor{red!70!black}{\textbf{(Incorrect retrieved theorem)}}
\end{tcolorbox}
\begin{tcolorbox}[
    colback=green!10,
    colframe=green!50!black,
    width=\textwidth,
    boxrule=0pt,
    arc=0mm,
    left=6pt,
    right=6pt,
    top=6pt,
    bottom=6pt,
    title=\textbf{\large \projectname{} GTE-Qwen2-7B-instruct (all query types) retriever},
    coltitle=black,
    fonttitle=\bfseries,
]
\textbf{Top Retrieved Theorem}: \textit{Ramsey's Theorem guarantees that for any edge-coloring of a sufficiently large complete graph, there exists a monochromatic complete subgraph. More formally, for integers $n_1, n_2, \ldots, n_c$, there exists a Ramsey number $R(n_1, \ldots, n_c)$ such that any $c$-coloring of a complete graph on $R(n_1, \ldots, n_c)$ vertices contains a monochromatic $K_{n_i}$ in some color $i$.}
\textcolor{green!40!black}{\textbf{(Correct theorem retrieved)}}
\end{tcolorbox}
\begin{tcolorbox}[
    colback=gray!10,
    colframe=orange!80!black,
    width=\textwidth,
    boxrule=0pt,
    arc=0mm,
    left=6pt,
    right=6pt,
    top=6pt,
    bottom=6pt,
    title=\textbf{\large Explanation},
    coltitle=black,
    fonttitle=\bfseries,
]
\textbf{Explanation:} While the Pigeonhole Principle might appear relevant at first glance due to the presence of element selection in the query, the problem is more appropriately modeled as a graph coloring task. Specifically, it reduces to a two-coloring of the edges of a complete graph, where the objective is to guarantee the existence of a monochromatic triangle. Our \projectname{} retriever demonstrates the ability to perform such nuanced reasoning, correctly interpreting the structural semantics of the query and retrieving the correct gold document: the Ramsey Theorem.
\end{tcolorbox}
\end{tcolorbox}
\caption{Example of \projectname{} success case compared to Qwen2, from TheoremQA theorems of \textsc{Bright}.}
\label{qualanalysis}
\end{figure*}

%% file: failure_cases.tex
\begin{figure*}[ht]
\begin{tcolorbox}[
    enhanced,
    sharp corners,
    colback=white,
    colframe=black,
    boxrule=0.8pt,
    width=\textwidth,
    breakable,
    left=0pt,
    right=0pt,
    top=0pt,
    bottom=0pt,
    boxsep=0pt,
    arc=0mm,
]
\begin{tcolorbox}[
    colback=blue!10,
    colframe=blue!60,
    width=\textwidth,
    boxrule=0pt,
    arc=0mm,
    left=6pt,
    right=6pt,
    top=6pt,
    bottom=6pt,
    title=\textbf{\large Example},
    coltitle=black,
    fonttitle=\bfseries,
]
\textbf{Query:} Imagine you have a family tree that shows the lineage from ancestors to their descendants, with arrows pointing from parents to children. Can this family tree, with its directed lineage paths, be accurately represented without the arrows while still maintaining the correct relationships? True or false? \\ \\
\noindent\textbf{Gold Theorem Id}: 3778 \\
\textbf{Gold Theorem}: \textit{The '''girth''' of $G$ is the smallest length of any cycle in $G$. An acyclic graph is defined as having a girth of infinity.}
\textit{(Note that the gold theorem has no lexical overlap with the original question.)}
\end{tcolorbox}
\begin{tcolorbox}[
    colback=red!10,
    colframe=red!60,
    width=\textwidth,
    boxrule=0pt,
    arc=0mm,
    left=6pt,
    right=6pt,
    top=6pt,
    bottom=6pt,
    title=\textbf{\large Qwen2 retriever},
    coltitle=black,
    fonttitle=\bfseries,
]
\textbf{Top Retrieved Theorem}: \textit{Rooted Tree Corresponds to Arborescence: Let $T = (V, E)$ be a rooted tree with root $r$. Then there exists a unique orientation of $T$ that forms an $r$-arborescence.}\\ \\
\textcolor{red!70!black}{\textbf{(Incorrect theorem retrieved)}}
\end{tcolorbox}
\begin{tcolorbox}[
    colback=green!10,
    colframe=green!50!black,
    width=\textwidth,
    boxrule=0pt,
    arc=0mm,
    left=6pt,
    right=6pt,
    top=6pt,
    bottom=6pt,
    title=\textbf{\large \projectname{} GTE-Qwen2-7B-instruct (all query types) retriever},
    coltitle=black,
    fonttitle=\bfseries,
]
\textbf{Top Retrieved Theorem}: \textit{Rooted Tree Corresponds to Arborescence: Let $T = (V, E)$ be a rooted tree with root $r$. Then there exists a unique orientation of $T$ that forms an $r$-arborescence.}\\ \\
\textcolor{red!70!black}{\textbf{(Incorrect theorem retrieved)}}
\end{tcolorbox}
\begin{tcolorbox}[
    colback=gray!10,
    colframe=orange!80!black,
    width=\textwidth,
    boxrule=0pt,
    arc=0mm,
    left=6pt,
    right=6pt,
    top=6pt,
    bottom=6pt,
    title=\textbf{\large Explanation},
    coltitle=black,
    fonttitle=\bfseries,
]
\textbf{Explanation:} This case demonstrates a semantic mismatch: while the retrieved theorem on arborescences aligns superficially with the query’s mention of family trees, the actual focus is on acyclicity, better captured by the gold theorem on girth. The error highlights how surface-level similarity can mislead retrieval.
\end{tcolorbox}
\end{tcolorbox}
\caption{Example of \projectname{} Failure cases from TheoremQA theorems BRIGHT}
\label{qualanalysis_failure}
\end{figure*}

%% file: codingexample.tex
\begin{figure*}[ht]
\begin{tcolorbox}[
    enhanced,
    sharp corners,
    colback=white,
    colframe=black,
    boxrule=0.8pt,
    width=\textwidth,
    breakable,
    left=0pt,
    right=0pt,
    top=0pt,
    bottom=0pt,
    boxsep=0pt,
    arc=0mm,
]
\begin{tcolorbox}[
    colback=blue!10,
    colframe=blue!60,
    width=\textwidth,
    boxrule=0pt,
    arc=0mm,
    left=6pt,
    right=6pt,
    top=6pt,
    bottom=6pt,
    title=\textbf{\large Example},
    coltitle=black,
    fonttitle=\bfseries,
]
\textbf{Query:}\\ \\ 
\textbf{Problem:} Given an array of integers \texttt{heights} representing the histogram's bar heights (each bar has width 1), return the area of the largest rectangle in the histogram.

\textbf{Example 1.} Input: \texttt{heights = [2,1,5,6,2,3]}. Output: \texttt{10}. The largest rectangle has area 10.

\textbf{Example 2.} Input: \texttt{heights = [2,4]}. Output: \texttt{4}.

\textbf{Constraints.} $1 \leq \texttt{heights.length} \leq 10^5$, $0 \leq \texttt{heights[i]} \leq 10^4$.
\end{tcolorbox}
\begin{tcolorbox}[
    colback=red!10,
    colframe=red!60,
    width=\textwidth,
    boxrule=0pt,
    arc=0mm,
    left=6pt,
    right=6pt,
    top=6pt,
    bottom=6pt,
    title=\textbf{\large BM25 retriever},
    coltitle=black,
    fonttitle=\bfseries,
]
\textbf{Top Retrieved Document}:\\ 
\textbf{Problem:} Given an array \texttt{rectangles}, where \texttt{rectangles[i] = [\textit{width}$_i$, \textit{height}$_i$]}, return the number of pairs $(i, j)$ with $i < j$ such that the rectangles have the same width-to-height ratio, i.e., $\frac{\textit{width}_i}{\textit{height}_i} = \frac{\textit{width}_j}{\textit{height}_j}$ (using decimal division).

\textcolor{red!70!black}{\textbf{(Incorrect retrieved theorem)}}
\end{tcolorbox}
\begin{tcolorbox}[
    colback=green!10,
    colframe=green!50!black,
    width=\textwidth,
    boxrule=0pt,
    arc=0mm,
    left=6pt,
    right=6pt,
    top=6pt,
    bottom=6pt,
    title=\textbf{\large \projectname{} Llama3.1-8B-instruct retriever},
    coltitle=black,
    fonttitle=\bfseries,
]
\textbf{Top Retrieved Document}:\\ 
\textbf{Problem:} Given a binary matrix \texttt{matrix} of size \texttt{rows} $\times$ \texttt{cols}, return the area of the largest rectangle containing only \texttt{'1'}s.\\
\textbf{Approach:} For each row, treat it as the base of a histogram where the height at each column counts consecutive \texttt{'1'}s up to that row. For each histogram, compute the largest rectangle using a stack-based approach (\textit{similar to the Largest Rectangle in Histogram problem}).

\textbf{Constraints.} $1 \leq \texttt{rows}, \texttt{cols} \leq 200$; $\texttt{matrix[i][j]} \in \{'0', '1'\}$.
\textcolor{green!30!black}{\textbf{(Correct theorem retrieved)}}
\end{tcolorbox}
\begin{tcolorbox}[
    colback=gray!10,
    colframe=orange!80!black,
    width=\textwidth,
    boxrule=0pt,
    arc=0mm,
    left=6pt,
    right=6pt,
    top=6pt,
    bottom=6pt,
    title=\textbf{\large Explanation},
    coltitle=black,
    fonttitle=\bfseries,
]
The coding problem has a mathematical substructure because it requires computing the maximum rectangular area under a histogram, which involves geometric reasoning and optimization over intervals of the graph. Such a structure helps \projectname{} to do better at code retrieval tasks.
\end{tcolorbox}
\end{tcolorbox}
\caption{Analysis of \projectname{} Llama3.1-8B-instruct success case for Coding Example from \textit{Leetcode} split of BRIGHT}
\label{codingexample}
\end{figure*}

%% file: Appendix_B.tex
\clearpage
\section{Example Prompts}
\label{sec:appendix_b}
This section provides the prompts used in our work.

\input{Prompts/prompt3}
\section{Prompt for OST thought action}
\label{OST:appendix}
We provide the prompts used for our \textbf{One Step Thought Action} (OST) $A_1$ used in the MCTS, in Figure~\ref{fig:OSTprompt}. 
\input{Prompts/prompt4}
\section{Prompt for CRS action}
\label{CRS:appendix}
We provide the prompts for the Complete Reasoning Steps action $A_2$ of MCTS, in Figure~\ref{fig:CRSprompt}.  
\input{Prompts/prompt1}
\input{Prompts/promtpt_fewshot_MCTSquerygen}
\section{Prompt for MCTS Query Generation}
\label{querygen:appendix}
In this section, we provide the prompts for the query generation action $A_3$ used in our MCTS framework. These prompts guide LLM$_\mathrm{gen}$ to generate a plausible hypothetical theorem that could help solve the current subproblem. This format of the prompt is chosen to increase the likelihood of term matching retrievers finding the relevant theorems. The instruction is provided in Figure~\ref{fig:querygenMCTSinst} and the few shot examples used are presented in Figure~\ref{fig:querygenMCTSfewshot}.
\input{Prompts/prompt2}
\input{Prompts/prompt5}
\section{Prompt for Self Reflection MCTS}
\label{selfreflection:appendix}
In this section, we provide the prompts for the \textit{self reflection} and \textit{self-summarization} mechanisms used in our MCTS framework. The self-reflection prompt is shown in Figure~\ref{fig:selfrefMCTS} and the  summatization prompt used is presented in Figure~\ref{prompt:selfsumm}.
\section{Prompt for LLM Reasoning Query generation}
\label{LLMquery:appendix}
In this section, we provide the prompts used for generating LLM$_\mathrm{query}$ based on the input math question, the reasoning CoT context and the theorem retrieved in the MCTS. The instruction with the few-shot examples are provided in Figure~\ref{fig:LLMquery}.

\input{Prompts/prompt6}

%% file: Prompts/prompt3.tex
\begin{figure*}[ht]
\begin{tcolorbox}[
    enhanced,
    sharp corners,
    colback=white,
    colframe=black,
    boxrule=0.8pt,
    width=\textwidth,
    breakable,
    left=0pt,
    right=0pt,
    top=0pt,
    bottom=0pt,
    boxsep=0pt,
    arc=0mm,
]
\begin{tcolorbox}[
    colback=blue!10,
    colframe=blue!60,
    width=\textwidth,
    boxrule=0pt,
    arc=0mm,
    left=6pt,
    right=6pt,
    top=6pt,
    bottom=6pt,
    title=\textbf{\large Instruction},
    coltitle=black,
    fonttitle=\bfseries,
]
A chat between a curious user and an AI assistant. The assistant gives step-by-step solutions to the user's questions. At the final step, a conclusive answer is given in the format of "The answer is: boxed\{ANSWER\}.", where ANSWER should be a numeric answer. \\

 \{\textit{few shot examples}\} \\
 \textbf{Question:} \{question\} \\
 \textbf{Response:} Let's think step by step.\\
\end{tcolorbox}
\begin{tcolorbox}[
    colback=red!10,
    colframe=red!60,
    width=\textwidth,
    boxrule=0pt,
    arc=0mm,
    left=6pt,
    right=6pt,
    top=6pt,
    bottom=6pt,
    title=\textbf{\large Few Shot Examples},
    coltitle=black,
    fonttitle=\bfseries,
]
\textbf{Question}: Gracie and Joe are choosing numbers on the complex plane. Joe chooses the point $1+2i$. Gracie chooses $-1+i$. How far apart are Gracie and Joe's points?\\
\textbf{Response}: Let's think step by step.\\
Step 1: Distance formula is $\sqrt{(x_2-x_1)^2+(y_2-y_1)^2}$.\\
Step 2: Joe's point $(1,2)$, Gracie's $(-1,1)$.\\
Step 3: Distance $\sqrt{((-1)-(1))^2+((1)-(2))^2}=\sqrt{4+1}=\sqrt{5}$.\\
Step 4: Points are $\boxed{\sqrt{5}}$ units apart.\\
Step 5: The answer is: $\sqrt{5}$.\\[5pt]

\textbf{Question}: What is the sum of all positive integer values of $n$ for which $\frac{n+6}{n}$ is an integer?\\
\textbf{Response}: Let's think step by step.\\
Step 1: $n+6$ divisible by $n$.\\
Step 2: $n$ must be factor of 6.\\
Step 3: Factors of 6 are 1, 2, 3, 6. Sum is $1+2+3+6=\boxed{12}$.\\
Step 4: The answer is: 12.\\[5pt]

\textbf{Question}: Abigail, Beatrice, and Carson sell eggs in cartons of 12. Abigail has 37, Beatrice 49, Carson 14 eggs. How many eggs remain after selling all cartons?\\
\textbf{Response}: Let's think step by step.\\
Step 1: Total eggs $37+49+14=100$.\\
Step 2: Divide by 12: $100\div12=8$ cartons, remainder 4.\\
Step 3: Remaining eggs $\boxed{4}$.\\
Step 4: The answer is: 4.\\[5pt]

\textbf{Question}: Circle $T$ has center $T(-2,6)$, reflected across $y$-axis, translated 8 units down. Find coordinates of image center.\\
\textbf{Response}: Let's think step by step.\\
Step 1: Reflect across $y$-axis: $(-(-2),6)=(2,6)$.\\
Step 2: Translate down 8 units: $(2,6-8)=(2,-2)$.\\
Step 3: Image coordinates $\boxed{(2,-2)}$.\\
Step 4: The answer is: $(2,-2)$.

\end{tcolorbox}
\end{tcolorbox}

\caption{Prompt for MCTS One Step Thought Action}
\label{fig:OSTprompt}
\end{figure*}

%% file: Prompts/prompt4.tex
\begin{figure*}[ht]
\begin{tcolorbox}[
    enhanced,
    sharp corners,
    colback=white,
    colframe=black,
    boxrule=0.8pt,
    width=\textwidth,
    breakable,
    left=0pt,
    right=0pt,
    top=0pt,
    bottom=0pt,
    boxsep=0pt,
    arc=0mm,
]
\begin{tcolorbox}[
    colback=blue!10,
    colframe=blue!60,
    width=\textwidth,
    boxrule=0pt,
    arc=0mm,
    left=6pt,
    right=6pt,
    top=6pt,
    bottom=6pt,
    title=\textbf{\large Instruction},
    coltitle=black,
    fonttitle=\bfseries,
]
You are given context about a mathematical question.  Your job is to generate the next steps of the solution and complete the solution. In the end of your response, a final answer is given in the format of  "\$\textbackslash boxed<ANSWER>\$" , where <ANSWER> should be a numeric result or a math expression.
\end{tcolorbox}
\begin{tcolorbox}[
    colback=red!10,
    colframe=red!60,
    width=\textwidth,
    boxrule=0pt,
    arc=0mm,
    left=6pt,
    right=6pt,
    top=6pt,
    bottom=6pt,
    title=\textbf{\large Few Shot Examples},
    coltitle=black,
    fonttitle=\bfseries,
]
\textbf{Context:} Gracie and Joe are choosing numbers on the complex plane. Joe chooses the point $1+2i$. Gracie chooses $-1+i$. How far apart are Gracie and Joe's points? The distance between two points $(x_1,y_1)$ and $(x_2,y_2)$ in the complex plane is given by $\sqrt{(x_2-x_1)^2+(y_2-y_1)^2}$. Joe's point is $(1,2)$ and Gracie's point is $(-1,1)$.\\
\textbf{Next steps:} The distance is $\sqrt{((-1)-(1))^2+((1)-(2))^2}=\sqrt{4+1}=\sqrt{5}$. Therefore, Gracie and Joe's points are $\boxed{\sqrt{5}}$ units apart.\\[5pt]

\textbf{Context:} What is the sum of all positive integer values of $n$ for which $\frac{n+6}{n}$ is an integer?\\
\textbf{Next steps:} We want $\frac{n+6}{n}$ to be integer, thus $n+6$ divisible by $n$. Since $n$ positive, $n$ must factor 6. Factors: 1, 2, 3, 6. Sum is $1+2+3+6=\boxed{12}$.\\[5pt]

\textbf{Context:} Abigail, Beatrice, and Carson sell eggs in cartons of 12. Abigail has 37 eggs, Beatrice has 49, Carson has 14. First, total eggs: $37+49+14=100$. Divide by 12: $100 \div 12 = 8$ remainder $4$.\\
\textbf{Next steps:} Eggs remaining: $\boxed{4}$. The answer is: 4.\\[5pt]

\textbf{Context:} Circle $T$ has center $T(-2,6)$, reflected across $y$-axis, translated 8 units down. Reflecting across $y$-axis negates $x$-coordinate.\\
\textbf{Next steps:} Reflection gives $(2,6)$. Translating down 8 units: $(2,6-8)=(2,-2)$. Therefore, coordinates are $\boxed{(2,-2)}$.

\end{tcolorbox}
\end{tcolorbox}

\caption{Prompt for Complete Remaining Steps (CRS) Action}
\label{fig:CRSprompt}
\end{figure*}

%% file: Prompts/prompt1.tex
\begin{figure*}[t]
\begin{tcolorbox}[
    enhanced,
    sharp corners,
    colback=white,
    colframe=black,
    boxrule=0.8pt,
    width=\textwidth,
    breakable,
    left=0pt,
    right=0pt,
    top=0pt,
    bottom=0pt,
    boxsep=0pt,
    arc=0mm,
]
\begin{tcolorbox}[
    colback=blue!10,
    colframe=blue!60,
    width=\textwidth,
    boxrule=0pt,
    arc=0mm,
    left=6pt,
    right=6pt,
    top=6pt,
    bottom=6pt,
    title=\textbf{\large Instruction},
    coltitle=black,
    fonttitle=\bfseries,
]
You are given a mathematical question and an intermediate solution. What are the mathematical concepts, theorems, formulas that would be useful for solving this question. Please provide the theorem name, followed by the theorem statement, followed by the preconditions in the theorem, and why the preconditions are satisfied in the question we have. Also mention which specific subjects in math this theorem corresponds to. List out as many number of theorems that are highly relevant to this question. Do not output the final solution. Do not generate theorems which are already present in the intermediate solution.
\end{tcolorbox}
\end{tcolorbox}

\caption{Instruction for MCTS Query Generation Action}
\label{fig:querygenMCTSinst}
\end{figure*}

%% file: Prompts/promtpt_fewshot_MCTSquerygen.tex
\begin{figure*}[ht]
\begin{tcolorbox}[
    enhanced,
    sharp corners,
    colback=white,
    colframe=black,
    boxrule=0.8pt,
    width=\textwidth,
    breakable,
    left=0pt,
    right=0pt,
    top=0pt,
    bottom=0pt,
    boxsep=0pt,
    arc=0mm,
]
\begin{tcolorbox}[
    colback=red!10,
    colframe=red!60,
    width=\textwidth,
    boxrule=0pt,
    arc=0mm,
    left=6pt,
    right=6pt,
    top=6pt,
    bottom=6pt,
    title=\textbf{\large Few Shot Examples},
    coltitle=black,
    fonttitle=\bfseries,
]
\textbf{Theorem}: Polynomial Division Algorithm\\
\textbf{Theorem Statement}: For any two polynomials P(z)(dividend) and D(z) (divisor), with $deg(P(z)) \geq deg (D(z))$, there exist unique polynomials Q(z) (quotient) and R(z) (remainder) such that: $P(z)=D(z)Q(z)+R(z)$.\\
\textbf{Question}: Find quotient of $\frac{3z^4 -4z^3 +5z^2 -11z +2}{2+3z}$.\\
\textbf{Intermediate Solution}: Apply polynomial long division.\\
\textbf{Query}: (Polynomial Division) $P(z)=D(z)Q(z)+R(z)$ if $\deg(P) \geq \deg(D)$.\\
\textbf{Preconditions Met}: $(1)  D(z)=3z+2\neq 0$, (2) $\deg(P)=4>\deg(D)=1$.\\
\textbf{Subject}: Algebra.\\[5pt]

\textbf{Theorem}: Principle of Inclusion-Exclusion\\
\textbf{Theorem Statement}: For any two finite sets A and B, the size of their union is given by:
$|A \cup B| = |A| + |B| - |A \cap B|$.\\
\textbf{Question}: Probability of palindrome (letters/digits) in plates, simplified as $\frac{m}{n}$, find $m+n$.\\
\textbf{Intermediate Solution}: Compute separately, combine using inclusion-exclusion.\\
\textbf{Query}: (Inclusion-Exclusion) $|A \cup B|=|A|+|B|-|A \cap B|$.\\
\textbf{Preconditions Met}: (1) Sets finite (letters/digits), potential overlap possible.\\
\textbf{Subject}: Combinatorics.\\[5pt]

\textbf{Theorem}: Basic Multiplication Principle\\
\textbf{Theorem Statement}: If there are m ways to do something and n ways to do another thing, then there are m*n to do both things.\\
\textbf{Question}: Pages written/year if 3-page letters to 2 friends twice weekly?\\
\textbf{Intermediate Solution}: Pages/week calculation, then yearly total.\\
\textbf{Query}: (Multiplication Principle) Actions with $m$ and $n$ ways yield $m\times n$ combined ways.\\
\textbf{Preconditions Met}: Counts defined clearly; independent actions.\\
\textbf{Subject}: Arithmetic.

\end{tcolorbox}
\end{tcolorbox}

\caption{Few shot examples for MCTS Query Generation Action}
\label{fig:querygenMCTSfewshot}
\end{figure*}

%% file: Prompts/prompt2.tex
\begin{figure*}[ht]
\begin{tcolorbox}[
    enhanced,
    sharp corners,
    colback=white,
    colframe=black,
    boxrule=0.8pt,
    width=\textwidth,
    breakable,
    left=0pt,
    right=0pt,
    top=0pt,
    bottom=0pt,
    boxsep=0pt,
    arc=0mm,
]
\begin{tcolorbox}[
    colback=blue!10,
    colframe=blue!60,
    width=\textwidth,
    boxrule=0pt,
    arc=0mm,
    left=6pt,
    right=6pt,
    top=6pt,
    bottom=6pt,
    title=\textbf{\large Instruction},
    coltitle=black,
    fonttitle=\bfseries,
]
You are given a mathematical question, an intermediate solution and a mathematical theorem which was retrieved denoted as Retrieved Document. First, please judge whether the mathematical theorem is relevant with the question and the intermediate solution, and put it in the relevant field. If the provided content is irrelevant to the question and the context, explain the reason in the relevant reason field. The format will be as follows:\\
\textbf{Question}: [question]\\
\textbf{Intermediate solution}: [intermediate solution]\\
\textbf{Retrieved Document}: [theorem]\\
\textbf{Relevant}: [relevance label]\\
\textbf{Reason}: [reason].
\end{tcolorbox}
\begin{tcolorbox}[
    colback=red!10,
    colframe=red!60,
    width=\textwidth,
    boxrule=0pt,
    arc=0mm,
    left=8pt,
    right=8pt,
    top=0pt,
    bottom=0pt,
    title=\textbf{\large Few Shot Examples},
    coltitle=black,
    fonttitle=\bfseries,
]
\textbf{Question}: Let $f: \mathbb{R} \to \mathbb{R}$ be continuous, with $f(0)=0$ and $f(x+y)=f(x)+f(y)+xy$. Find degree of $f$.\\
\textbf{Intermediate Solution}: Define $g(x)=f(x)-\frac{x^2}{2}$.\\
\textbf{Retrieved Document}: Cauchy’s equation $f(x+y)=f(x)+f(y)$, continuous solution linear: $f(x)=cx$.\\
\textbf{Relevant}: True\\
\textbf{Reason}: Defining $g(x)$ transforms into Cauchy’s form, yielding $g(x)=cx$ and thus $f(x)=\frac{x^2}{2}+cx$ degree 2.\\[2pt]

\textbf{Question}: Let $f: \mathbb{R} \to \mathbb{R}$ differentiable with $f'(x)=f(x)+x$. Find $f(x)$.\\
\textbf{Intermediate Solution}: Requires solving differential equation directly.\\
\textbf{Retrieved Document}: Rolle’s Theorem guarantees $f'(c)=0$ under certain continuity/differentiability conditions.\\
\textbf{Relevant}: False\\
\textbf{Reason}: There is no direct application of Rolle's Theorem to the differential equation provided, as the theorem does not help in finding the solution to the equation $f'(x) = f(x) + x$.\\[2pt]

\textbf{Question}: Battery lifetime exponential mean 10 hours. Probability battery lasts at least 15 hours?\\
\textbf{Intermediate Solution}: Use exponential distribution's CDF.\\
\textbf{Retrieved Document}: Markov’s Inequality provides upper bound $P(X \geq a) \leq \frac{E[X]}{a}$.\\
\textbf{Relevant}: False\\
\textbf{Reason}: Markov’s gives bounds, not exact probabilities; exact CDF calculation is necessary here.\\[2pt]



\end{tcolorbox}
\end{tcolorbox}

\caption{Prompt for Self Reflection MCTS}
\label{fig:selfrefMCTS}
\end{figure*}

%% file: Prompts/prompt5.tex
\begin{figure*}[ht]
\begin{tcolorbox}[
    enhanced,
    sharp corners,
    colback=white,
    colframe=black,
    boxrule=0.8pt,
    width=\textwidth,
    breakable,
    left=0pt,
    right=0pt,
    top=0pt,
    bottom=0pt,
    boxsep=0pt,
    arc=0mm,
]
\begin{tcolorbox}[
    colback=blue!10,
    colframe=blue!60,
    width=\textwidth,
    boxrule=0pt,
    arc=0mm,
    left=6pt,
    right=6pt,
    top=6pt,
    bottom=6pt,
    title=\textbf{\large Instruction},
    coltitle=black,
    fonttitle=\bfseries,
]
Given the statement of a mathematical theorem in a structured latex format, convert it to a simpler natural language format by removing latex notations.
\end{tcolorbox}
\begin{tcolorbox}[
    colback=red!10,
    colframe=red!60,
    width=\textwidth,
    boxrule=0pt,
    arc=0mm,
    left=6pt,
    right=6pt,
    top=6pt,
    bottom=6pt,
    title=\textbf{\large Few Shot Examples},
    coltitle=black,
    fonttitle=\bfseries,
]
\textbf{Input ProofWiki Theorem (in latex)}:\\
<\textit{Latex Section}>: Quadratic Irrational is Root of Quadratic Equation \\ 
<\textit{Tags}>: Algebra, Quadratic Equations, Quadratic Irrationals \\
<\textit{begin theorem}> Let $x$ be a quadratic irrational. Then $x$ is a solution to a quadratic equation with rational coefficients.\\ \\ 
\textbf{Generated Natural language theorem}: Quadratic Irrational is Root of Quadratic Equation - A quadratic irrational number is always the root of some quadratic equation with rational coefficients.

\end{tcolorbox}
\end{tcolorbox}

\caption{Prompt for Self Summarization of Retrieved theorems}
\label{prompt:selfsumm}
\end{figure*}

%% file: Prompts/prompt6.tex
\clearpage
\begin{figure*}[ht]
\begin{tcolorbox}[
    enhanced,
    sharp corners,
    colback=white,
    colframe=black,
    boxrule=0.8pt,
    width=\textwidth,
    breakable,
    left=0pt,
    right=0pt,
    top=0pt,
    bottom=0pt,
    boxsep=0pt,
    arc=0mm,
]
\begin{tcolorbox}[
    colback=blue!10,
    colframe=blue!60,
    width=\textwidth,
    boxrule=0pt,
    arc=0mm,
    left=6pt,
    right=6pt,
    top=6pt,
    bottom=6pt,
    title=\textbf{\large Instruction},
    coltitle=black,
    fonttitle=\bfseries,
]
Given a math question, its partial solution (may be empty), and a retrieved theorem, do the following: Identify the preconditions of the theorem and explain why they hold in the given question. Using these preconditions, generate a general retrieval query that captures the key mathematical idea needed in the partial solution. The query should be a single sentence.
\end{tcolorbox}
\begin{tcolorbox}[
    colback=red!10,
    colframe=red!60,
    width=\textwidth,
    boxrule=0pt,
    arc=0mm,
    left=10pt,
    right=10pt,
    top=0pt,
    bottom=0pt,
    title=\textbf{\large Few Shot Examples},
    coltitle=black,
    fonttitle=\bfseries,
]
\textbf{Question}: A warehouse needs to store 65 identical boxes using a set of identical shelves, each of which can hold up to 8 boxes. What is the minimum number of shelves required to store all the boxes?\\
\textbf{Partial solution}: To determine the minimum number of shelves required, we divide the total number of boxes by the capacity of each shelf. Since the number of shelves must be a whole number, we round up to 9 shelves.\\
\textbf{Theorem}: If $n$ items are put into $m$ containers, with $n > m$, then at least one container must contain more than one item.\\
\textbf{Preconditions}: (1) There are more items than containers. (2) The items are distributed into containers.\\
\textbf{Why Preconditions are Satisfied:} (1) The warehouse has 65 boxes (items) and needs to distribute them among shelves (containers), where each shelf can hold up to 8 boxes. (2) Since 65 is greater than 8, multiple boxes must be placed on each shelf to store all of them.\\ 
\textbf{\textcolor{brown!40!black}{Generated Query:}} Minimizing the number of boxes needed to store a given number of objects with fixed capacity constraints.\\
\end{tcolorbox}
\end{tcolorbox}

\caption{Prompt for LLM generated query generation}
\label{fig:LLMquery}
\end{figure*}